\documentclass[sigconf]{acmart}
\usepackage{graphicx}
\usepackage{subfigure}
\usepackage{multirow}
\usepackage{bm}
\usepackage{float}
\usepackage[linesnumbered,commentsnumbered,ruled,vlined]{algorithm2e}

\AtBeginDocument{%
  \providecommand\BibTeX{{%
    \normalfont B\kern-0.5em{\scshape i\kern-0.25em b}\kern-0.8em\TeX}}}

\begin{document}
\copyrightyear{2020}
\acmYear{2020}
\setcopyright{acmcopyright}
\acmConference[SIGSPATIAL '20]{28th International Conference on Advances in Geographic Information Systems}{November 3--6, 2020}{Seattle, WA, USA}
\acmBooktitle{28th International Conference on Advances in Geographic Information Systems (SIGSPATIAL '20), November 3--6, 2020, Seattle, WA, USA}
\acmPrice{15.00}
\acmDOI{10.1145/3397536.3422268}
\acmISBN{978-1-4503-8019-5/20/11}

\title{
Reimagining City Configuration: Automated Urban Planning via Adversarial Learning
}

\author{Dongjie Wang}
\email{wangdongjie@Knights.ucf.edu}
\affiliation{%
  \institution{University of Central Florida}
  \streetaddress{}
  \city{Orlando}
  \state{Florida}
  \country{United States}
  \postcode{}
}

\author{Yanjie Fu}
\authornote{Concat Author.}
\email{yanjie.fu@ucf.edu}
\affiliation{%
  \institution{University of Central Florida}
  \streetaddress{}
  \city{Orlando}
  \state{Florida}
  \country{United States}
  \postcode{}
}

\author{Pengyang Wang}
\email{pengyang.wang@knights.ucf.edu}
\affiliation{%
  \institution{University of Central Florida}
  \streetaddress{}
  \city{Orlando}
  \state{Florida}
  \country{United States}
  \postcode{}
}

\author{Bo Huang}
\email{bohuang@cuhk.edu.hk}
\affiliation{%
  \institution{The Chinese University of Hong Kong}
  \streetaddress{}
  \city{Hong Kong}
  \country{China}
  \postcode{}
}

\author{Chang-Tien Lu}
\email{ctlu@vt.edu}
\affiliation{%
  \institution{Virginia Tech}
  \streetaddress{}
  \city{}
  \state{Virginia}
  \country{United States}
  \postcode{}
}



\renewcommand{\shortauthors}{Wang, et al.}

\begin{abstract}
Urban planning refers to the efforts of designing land-use configurations. 
Effective urban planning can help to mitigate the operational and social vulnerability of a urban system, such as high tax, crimes, traffic congestion and accidents, pollution, depression, and anxiety. 
Due to the high complexity of urban systems, such tasks are mostly completed by  professional planners. But, human planners take longer time.
The recent advance of deep learning motivates us to ask: can machines learn at a human capability to automatically and quickly calculate land-use configuration, so human planners can finally adjust machine-generated plans for specific needs?
To this end, we formulate the automated urban planning problem into a task of learning to configure land-uses, given the surrounding spatial contexts. 
To set up the task, we define a land-use configuration as a longitude-latitude-channel tensor, where each channel is a category of POIs and the value of an entry  is the number of POIs. 
The objective is then to propose an adversarial learning framework that can automatically generate such tensor for an unplanned area.
In particular, we first characterize the contexts of surrounding areas of an unplanned area by learning representations from spatial graphs using geographic and human mobility data.
Second, we combine each unplanned area and its surrounding context representation as a tuple, and categorize all the tuples into positive  (well-planned areas) and negative samples (poorly-planned areas).
Third, we develop an adversarial land-use configuration approach, where the surrounding context representation is fed into a generator to generate a land-use configuration, and a discriminator learns to distinguish among positive and negative samples. 
Finally, we devise two new measurements to evaluate the quality of land-use configurations and present extensive experiment and visualization results to demonstrate the effectiveness of our method.

\end{abstract}




\maketitle

\section{Introduction}
Urban planning is an interdisciplinary and complex process that involves with public policy, social science, engineering, architecture, landscape, and other related field. 
In this paper, we refer urban planning to the efforts of designing land-use configurations, which is the reduced yet essential task of urban planning. 
Effective urban planning can help to mitigate the operational and social vulnerability of a urban system, such as high tax, crimes, traffic congestion and accidents, pollution, depression, and anxiety.


Due to the high complexity of urban systems, such planning tasks are mostly completed by professional planners. But, human planners take longer time.
The recent advance of deep learning, particularly deep adversarial learning, provide a great potential for teaching a machine to imagine and create. 
This observation motivates us to rethink urban planning in the era of artificial intelligence: What roles does deep learning play in urban planning? Can machines develop and learn at a human capability to automatically and quickly calculate land-use configurations? In this way, machines can be planning assistants and human planners can finally adjust machine-generated plans for specific needs.

All of the above evidence has shown that it is appealing to develop a data-driven AI-enabled automated urban planner. 
However, three unique challenges arise to achieve the goal: 
(1) How can we quantify a land-use configuration plan? 
(2) How can we develop a machine learning framework that can learn the good and the bad of existing urban communities in terms of land-use configuration policies? 
(3) How can we evaluate the quality of generated land-use configurations? 
Next, we will introduce our research insights and solutions to address the challenges.


First, as we aim to teach a machine to reimagine the land-use configuration of an area, it is naturally critical to define a machine-perceivable structure for a land-use configuration. 
In practice, the land-use configuration plan of a given area is visually defined by a set of Point of Interests (POIs) and their corresponding locations (e.g., latitudes and longitudes) and urban functionality categories (e.g., shopping, banks, education, entertainment, residential). 
A close look into such visually-perceived land-use configuration can reveal that the land-use configuration is indeed a high-dimensional indicator that illustrates what and where we should put into an unplanned area. 
There is not just location-location statistical autocorrelation but also location-functionality statistical autocorrelation in a land-use configuration. To capture such statistical correlations, we propose to represent a land use configuration plan as a latitude-longitude-channel tensor, where each channel is a specific category of POIs that are distributed across the unplanned area, and the value of an entry in the tensor is the number of POIs.  
In this way, the tensor can describe not just the location-location interaction of POIs, but also location-function interaction of POIs.

Second, after we define the quantitative version of a land-use configuration, our second question is that how we can teach a machine to automatically generate a land-use configuration?
We analyze large-scale urban residential community data, and identify an important observation: 
1) an urban community can be viewed as an attributed node in a socioeconomic network (city), and this node proactively interacts with surrounding nodes (environments);
2) the coupling, interaction, and coordination of a community and surrounding environments significantly influence the livability, vibrancy, and quality of a community. 
Based on this observation, we propose to  convert the land-use configuration planning problem as a new objective: to teach a machine to generate a land-use configuration tensor given the surrounding context/area.
In other words, the problem is reduced into the objective of learning a conditional probability function that maps a surrounding context representation to a well-planned configuration tensor, instead of a poorly-planned configuration tensor. 
The recently emerging deep adversarial learning provides a great potential to address the reduced objective. 
We reformulate the task into an adversarial learning paradigm, in which: 
1) A neural generator is analogized as a machine planner that generates a land-use configuration; 
2) The generator generates a configuration in terms of a pattern feature representation of surrounding spatial contexts; 
3) The surrounding context representation is learned via self-supervised representation learning collectively from spatial graphs. 
4) A neural discriminator is to classify whether the generated land-use configuration is well-planned (positive) or poorly-planned (negative). 
5) A new mini-max loss function is constructed to guide the generator to learn from the goods of well-planned areas and the bads of poorly-planned areas. 

Third, it is traditionally a very challenging open question to evaluate the quality of a generated land-use configuration.  
The most solid and sound validation is to collaborate with urban developer and city governments to implement an AI-generated configuration into an unplanned area, and observe the development of the area in the following years. 
However, this is not realistic. 
In this paper, we exploit two strategies to assess the generated configurations: 
1) We build one scoring model that outputs the score of a configuration by learning from training data.
Specifically, we utilize a machine learning model to learn the data distribution of the original land-use reconfiguration samples. 
Thus, after we obtain the generated solutions, the scoring model has the ability to give a score.
2) We invite experienced regional planning experts to evaluate the quality of the generated solutions. In this paper, we let human experts to perform analysis on multiple case studies. 

In summary, in this paper, we develop an adversarial learning framework to  generate effective land-use configurations by learning from urban geography, human mobility, and socioeconomic data. 
Specifically, our contributions are:
1) We develop a latitude-longitude-channel tensor to quantify a land-use configuration plan. 
2) We propose a socioeconomic interaction perspective to understand urban planning as a process of optimizing the coupling between a community and surrounding environments. 
3) We reformulate the automated urban planning problem into an adversarial learning framework that maps surrounding spatial contexts into a configuration tensor by a machine generator. 
4) Although evaluation is challenging, we develop multiple aspects to conduct extensive experiment and visualization with real-world data to show the value of our method.

\section{Problem Statement and Framework Overview}

\subsection{Definitions}
\subsubsection{Central area and its Contexts}
In this paper, a central area is a square area that is centered on a geographical location (i.e., latitude and longitude), where is an unplanned area. 
In this study, the area of a central area is $1km^2$. 
The contexts of a central area wrap the residential community from different directions.

\begin{figure}[htbp]
    \centering
    \includegraphics[width=0.35\textwidth]{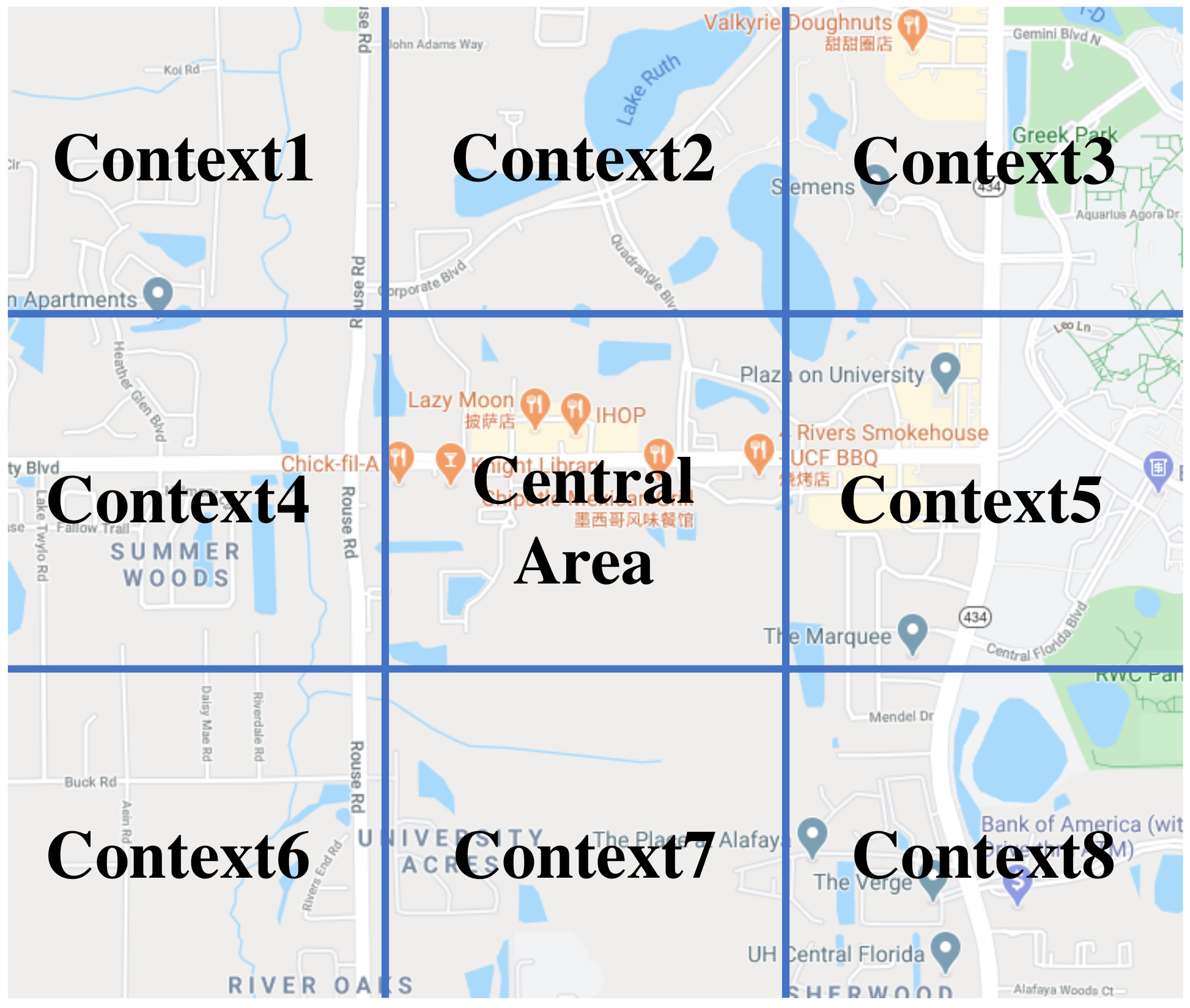}
    \caption{The geographical definitions of a central area and its surrounding spatial contexts.}
    \label{residential_context}
    \vspace{-0.5cm}
\end{figure}

Figure \ref{residential_context} represents the spatial relationship between a central area and its contexts.
In addition, we find that there are many POIs in the central area and its contexts.
Intuitively, the future development of an unplanned area is affected by its contexts.

\subsection{Problem Statement}
Urban planning is a complex field.
An excellent urban planning solution requires experts who own the amount of specific urban planning knowledge and experiences to spend much time designing.
In order to reduce the heavy burden of the experts and generate suitable urban planning solutions objectively, we propose an automatic urban planner that produces excellent solutions based on the context environments.  
Here, we simplify the meaning of urban planning into land-use configuration, which makes the question easier to model.
Formally, assuming a virgin area is $R$, the contexts of $R$ are $[C_1 \sim C_8]$, and the land-use configuration plan is $\mathbf{M}$ that structure is a multi-channel image, one channel represents one POI category data distribution.
Given the explicit feature $\mathbf{F}$ that describes the situation of the context environments, where the matrix $\mathbf{F} \in \mathbb{R}^{8 \times K}$, $8$ is the number of contexts and $K$ is the dimension of the explicit feature vector of each context.

\begin{figure}[!thbp]
    \centering
    \vspace{-0.3cm}
    \includegraphics[width=0.8\linewidth]{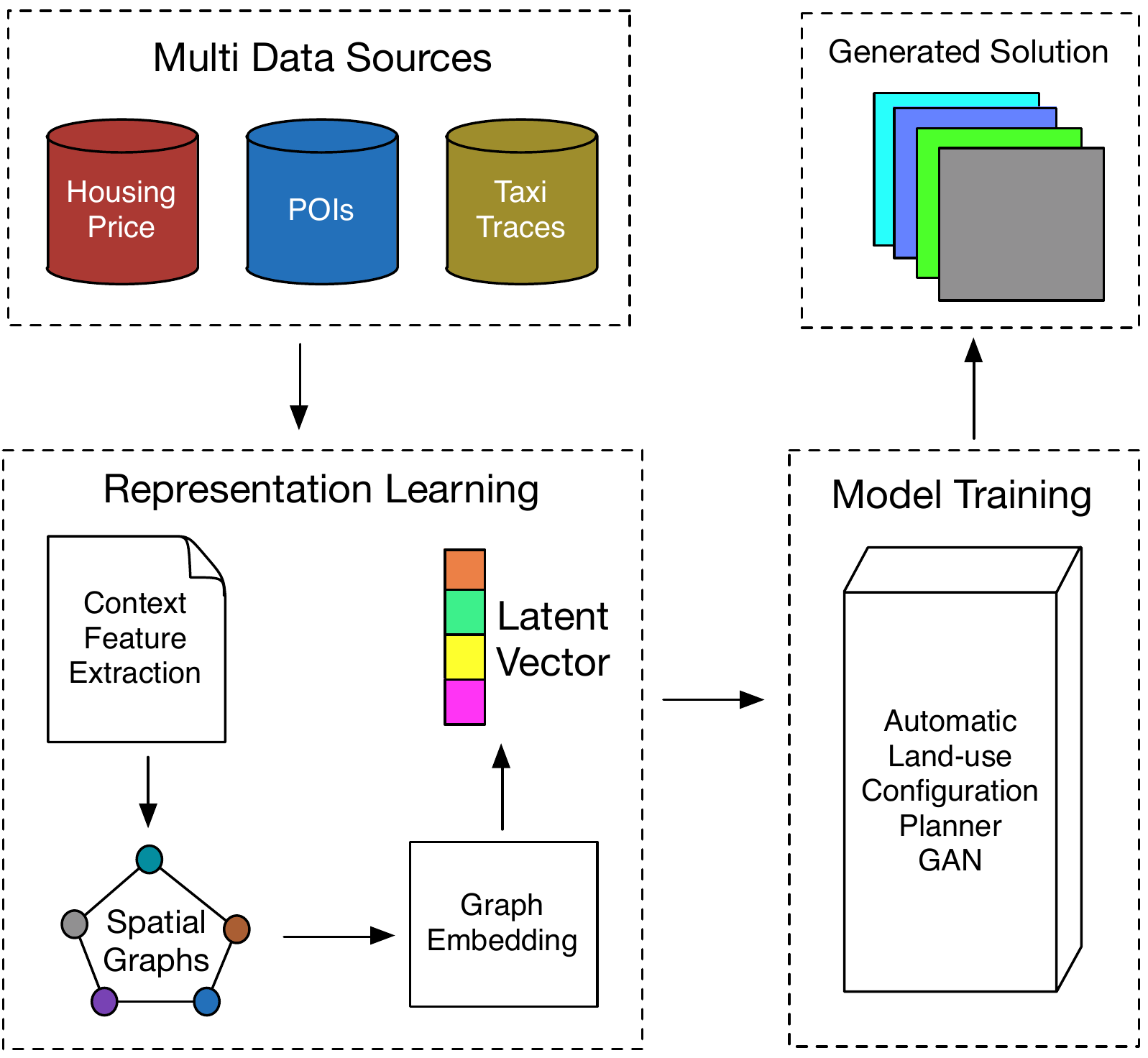}
    \vspace{-0.2cm}
    \caption{An overview of the proposed framework. The proposed framework includes four steps: 1) we first collect multiple data sources such as urban community related data (housing prices), point of interests data, and human mobility data (taxicab GPS traces). We then propose a spatial graph representation learning to learn the representation of surrounding contexts. Later, we develop an adversarial land use configuration machine to automate planning and generate recommended configurations. }
    \label{framework}
    \vspace{-0.5cm}
\end{figure}
The purpose of our framework is to take the explicit feature vector $\mathbf{F}$ as input and output the corresponding excellent land-use configuration solution $\mathbf{M}$.

\subsection{Framework Overview}

Figure \ref{framework} shows an overview of our proposed method (LUCGAN).
This framework has two main parts: 
(i) learning representation of the contexts of the virgin area;
(ii) generating an excellent land-use configuration solution for the virgin area.
In the first part, we first extract explicit features of the contexts from value-added space, POI distribution, public and private transportation conditions. 
Then, we construct a graph structure to capture the geographical spatial relationship between the virgin area and its contexts.
Afterward, we map the explicit features of contexts to the graph as attributes of corresponding nodes.
The attributed spatial graph incorporates all characteristics of contexts together.
Next, we utilize variational graph auto-encoder (VGAE) to obtain the latent representation of the contexts.
Thus, we get the final representation of the contexts of virgin areas through the first part.  
In the second part, we input the latent representation of the contexts, excellent land-use configuration samples, and terrible land-use configuration samples into an extended GAN.
The extended GAN is capable to generate the land-use configuration solution based on the contexts embedding.
Moreover, we customize a new GAN loss which makes the model learns the distribution of excellent plans and keeps away from the terrible plans.
Finally, when the model converges, the generator of the extended GAN can produce suitable and excellent land-use configuration solutions in an objective angle based on the latent context embedding.

\section{Automatic planner for land-use configuration}

In this section, we first introduce how to quantify surrounding context/area.
Then, we detail the measurement of how to evaluate the quality of land-use configuration solutions.
Finally, we describe how to train a generative model to learn an automated urban planner.


\subsection{Explicit Feature Extraction for Context Environments}
The land-use configuration solution of an unplanned area has a strong relationship with its contexts.
For example, if there are many commercial zones in the contexts of the unplanned area, we should avoid the redundancy of the same category POI in the planning. This is because we can make the unplanned area owns different function compared with its contexts, which is beneficial for the development and communication among the virgin area and its contexts.
Thus we grab the intrinsic characteristics of the contexts completely by extracting multiple explicit features.  

There are lots of indicators to describe context environments. 
Here, we select four views to capture the features of the contexts:

\begin{enumerate}
    \item \textbf{Value-added Space.}
        In common, the variation of house price reflects the value-added space of one area.
        Thus, we calculate the changing trend of house price of the contexts $[C_1 \sim C_8]$ in continued six months.
         Here, we take the context $C_1$ as an example to explain the calculation process.
         First, We obtain the housing price list among $t$ months.
         Then, we calculate the changing trend of house price by using the current house price value subtract the previous house price value.
         So we get the changing trend of $C_1$ as $\mathbf{v}_1=[v_1^1,v^2_1,...,v^{t-1}_1]$, where $v_1^i$ represents the value of the changing trend at i-th month.
         Finally, we collect the house price changing trend of all contexts together. The collected result is denoted as $\mathbf{V} = [\mathbf{v}_1,\mathbf{v}_2,...,\mathbf{v}_8]$, where the matrix $\mathbf{V} \in \mathbb{R}^{8 \times {t-1}}$.
         
    \item \textbf{POI Ratio.}
        Since various POIs provide diverse services to residents, the ratio of different types of POIs is a good indicator for indicating the functions of the area.
        Therefore, we calculate the POI ratio of the contexts $[C_1 \sim C_8]$.
        Here, we take $C_1$ as an example to explain the calculation process.
        First, we sum up the count of each POI category to form a feature vector.
        Then, we divide each item in the feature vector by the sum of all POI categories.
        We obtain the POI ratio of $C_1$, denoted by $\mathbf{r}_1 = [r_1^1,r^2_1,...,r^m_1]$, where $r_1^i$ represents the ratio of i-th POI category in $C_1$ and $m$ is the total number of POI categories.
        Finally, we collect the POI ratio of all contexts together.
        The collected result is denoted as $\mathbf{R} = [\mathbf{r}_1,\mathbf{r}_2,...,\mathbf{r}_8]$, where the matrix $\mathbf{R} \in \mathbb{R}^{8\times{m}}$, $m$ is the number of POI categories.
        
    \item \textbf{Public Transportation.}
    Public transportation is one popular travel mode due to its convenience and cheapness.
      So public transportation is a vital factor to be considered to describe the human mobility patterns.
      Thus, we extract features that related to public transportation to describe the public traffic situation of the contexts $C_1 \sim C_8$.
      We take $C_1$ as an example to show the calculation details.
      We calculate the feature vector of public transportation from five perspectives:
      (1) the leaving volume of $C_1$ in one day, denoted by $\mathbf{o}_1^1$;
      (2) the arriving volume of $C_1$ in one day, denoted by $\mathbf{o}_1^2$;
      (3) the transition volume of $C_1$ in one day, denoted by $\mathbf{o}_1^3$;
      (4) the density of bus stop of $C_1$, denoted by $\mathbf{o}_1^4$, which reflects the number of bus stop in per square meter;
      (5) the average balance of smart card of $C_1$, denoted by $\mathbf{o}_1^5$, which shows the economic expenditure of people in the travel field.
      The public transportation feature vector of $C_1$ can be denoted as $[o_1^1,o^2_1,...,o^5_1]$.
      Finally, we collect the public transportation feature vectors of all contexts together.
      The collected result is denoted as $\mathbf{O} = [\mathbf{o}_1,\mathbf{o}_2,...,\mathbf{o}_8]$, where the matrix $\mathbf{O} \in \mathbb{R}^{8 \times 5}$. 
     
     \item \textbf{Private Transportation.}
    Taxi is another important tool for people traveling.
     The taxi trajectory data reflects the people's flow count and the traffic congestion situation of an area.
     Thus, we explore the features of the private transportation condition of the contexts $[C_1 \sim C_8]$.
     Here, we take $C_1$ as an example to illustrate the calculation process.
     We count the features of private transportation from the following 5 perspectives:
     (1) the leaving volume of $C_1$ in one day, denoted by $u_1^1$;
     (2) the arriving volume of $C_1$ in one day, denoted by $u_1^2$;
     (3) the transition volume of $C_1$ in one day, denoted by $u_1^3$;
     (4) in $C_1$, the average driving velocity of a taxi in one hour, denoted by $u_1^4$;
     (5) in $C_1$, the average commute distance for a taxi, denoted by $u_1^5$;
     Then, the feature vector of private transportation can be denoted as $[u_1^1,u^2_1,...,u^5_1]$.
     Ultimately, we collect the private transportation feature vectors of all contexts together.
     The collected result is denoted as $\mathbf{U}=[\mathbf{u}_1,\mathbf{u}_2,...,\mathbf{u}_8]$, where the matrix $\mathbf{U} \in \mathbb{R}^{8 \times 5}$.
      
\end{enumerate}

After that, we obtain the explicit feature set of the contexts $C_1 \sim C_8$.
The set contains four kinds of features  $[\mathbf{V},\mathbf{R},\mathbf{O},\mathbf{U}]$, which describe the context environments from four perspectives.

\subsection{Explicit Features as Node Attributes: Constructing the Spatial Attributed Graph}

\begin{figure}[htbp]
    \centering
    \vspace{-0.3cm}
    \includegraphics[width=0.45\linewidth]{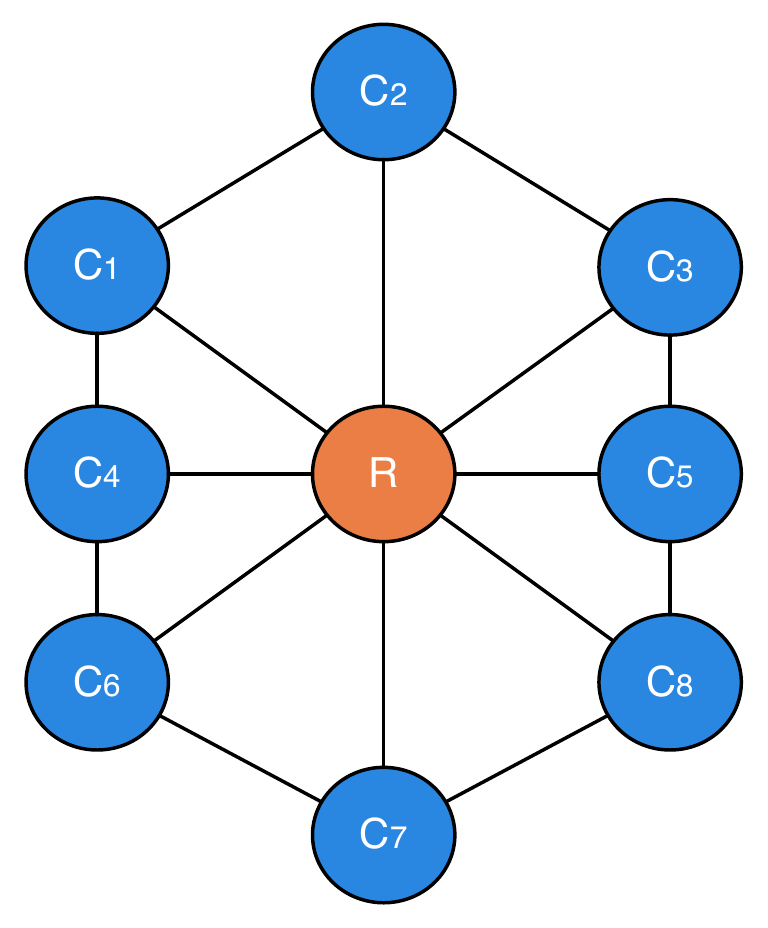}
        \vspace{-0.1cm}
    \caption{The graph structure between residential community and its surrounding spatial contexts. }
    \vspace{-0.4cm}
    \label{spatial_graph}
\end{figure}

The context environments wrap the residential community from different directions, resulting in spatial correlation among areas.
Such phenomenon motivates us to exploit spatial graphs to capture such spatial correlations.
Specifically, Figure \ref{spatial_graph} shows the graph structure, where the blue nodes represent the contexts;
the orange node is the residential community;
the edge between two nodes reflects the connectivity between them.

\begin{figure}[htbp]
    \centering
    \vspace{-0.3cm}
    \includegraphics[width=\linewidth]{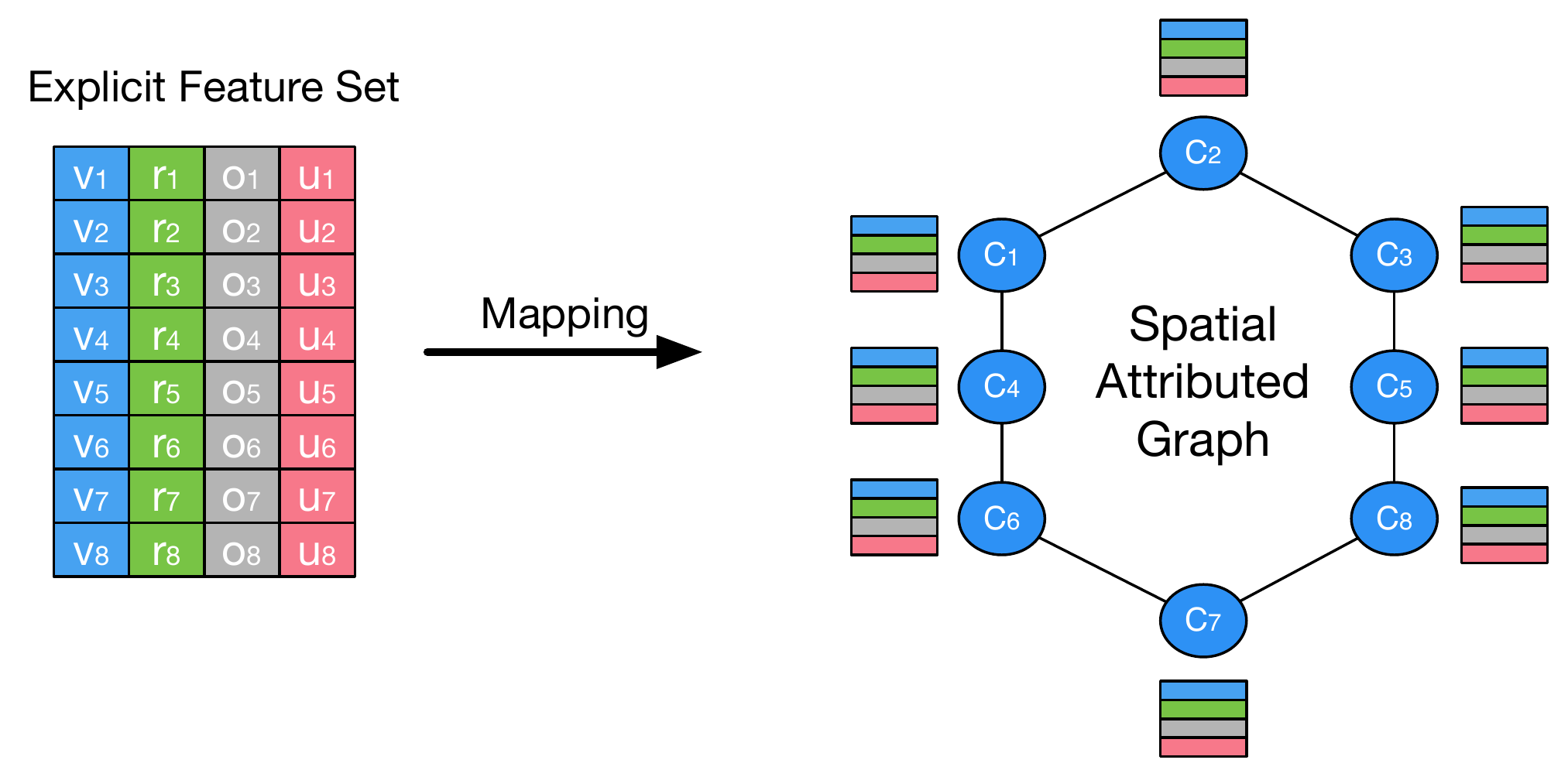}
    \vspace{-0.1cm}
    \caption{The illustration of constructing spatial attributed graphs: Each feature vector is mapped to the corresponding nodes by a column-wise strategy.}
    \vspace{-0.5cm}
    \label{graph_feature}
\end{figure}

Then, in order to fuse the spatial relationship and explicit features of the contexts, we construct a spatial attributed graph structure.
Formally, we map the explicit features to the spatial graph based on the corresponding context node as the node attribute.
Figure \ref{graph_feature} expresses the construction process of the spatial attributed graph.
This graph not only contains the explicit feature vector of the contexts but also includes the spatial relations among them.

\subsection{Learning Representation of the Spatial Attributed Graph}

\begin{figure}[t]
    \centering
    \vspace{-0.3cm}
    \includegraphics[width=\linewidth]{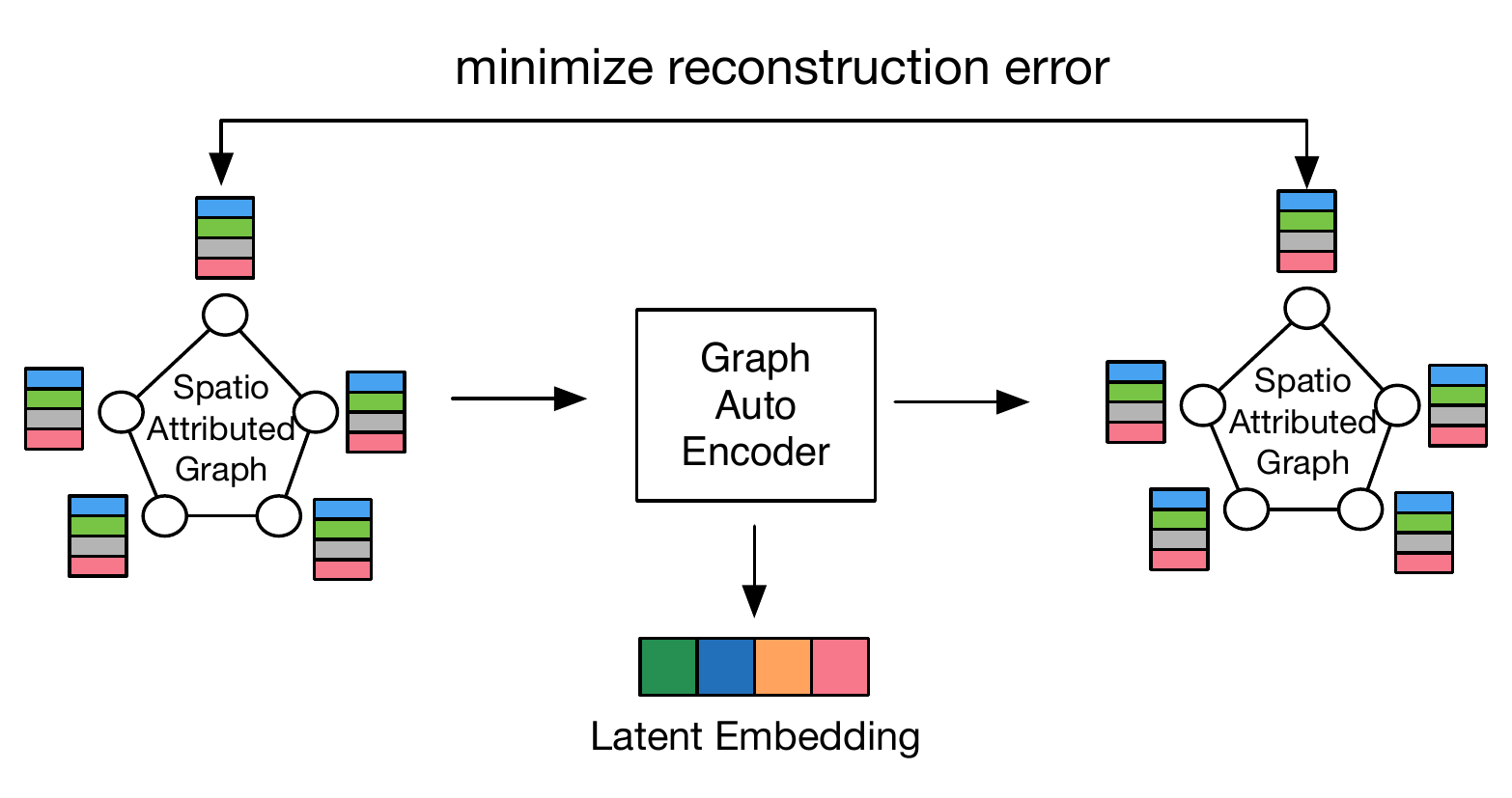}
    \vspace{-0.6cm}
    \caption{The proposed representation learning model to obtain surrounding  context representations by minimizing the reconstruction loss of spatial attributed graphs.}
    \vspace{-0.6cm}
    \label{gae_feature}
\end{figure}

Figure \ref{gae_feature} shows that we develop a spatial representation learning framework to preserve and fuse the explicit features and spatial relationship in the contexts. 
Formally, we denote the spatial attributed graph $G$ by $G=(\mathbf{X},\mathbf{A})$,  where $\mathbf{A}$ is the adjacency matrix that expresses the accessibility among different nodes;
$X$ is the feature matrix of the graph, here, $\mathbf{X}=[\mathbf{V},\mathbf{R},\mathbf{O},\mathbf{U}]$, and the concatenation direction is row-wise. 
In order to get the latent graph embedding $\mathbf{z}$,
we minimize the reconstruction loss between original graph $G$ and the reconstructed graph $\widehat{G}$ by an encoding-decoding framework.

The encoder part owns two Graph Convolutional Network (GCN) layers.
The first GCN layer takes $\mathbf{X}$ and $\mathbf{A}$ as input and output the feature matrix of low-dimensional space $\widehat{\mathbf{X}}$.
Thus, the encoding process can be formulated as:
\begin{equation}
    \widehat{\mathbf{X}} = GCN_1(\mathbf{X},\mathbf{A})=RELU(\widehat{\mathbf{D}}^{-\frac{1}{2}}\mathbf{A}\widehat{\mathbf{D}}^{-\frac{1}{2}}\mathbf{XW}_{1})
\end{equation}
where $\widehat{\mathbf{D}}$ is the diagonal degree matrix, $\mathbf{W}_1$ is the weight matrix of the $GCN_1$, and the whole layer is activated by $RELU$ function.

Owing to the latent embedding $\mathbf{z}$ is sampled from a prior Normal Distribution, the second GCN layer is responsible for assessing the parameters of the prior distribution.
Formally, the second GCN layer takes $\widehat{\mathbf{X}}$ and $\mathbf{A}$ as input and then outputs the mean value $\bm{\mu}$ and the variance value $\bm{\delta}^2$.
So the calcuation process of the second GCN layer can be formulated as:
\begin{equation}
    \bm{\mu},log(\bm{\delta}^2) = GCN_2(\mathbf{X},\mathbf{A}) = \widehat{\mathbf{D}}^{-\frac{1}{2}}\mathbf{A}
    \widehat{\mathbf{D}}^{-\frac{1}{2}}\widehat{\mathbf{X}}\mathbf{W}_2
\end{equation}
where $\mathbf{W}_2$ is the weight matrix of $GCN_2$.
Next, we use the reparameterization trick to approximate the sample operation to obtain the latent representation $\mathbf{z}$:

\begin{equation}
    \mathbf{z}=\bm{\mu}+\bm{\delta} \times \epsilon
\end{equation}
where $\epsilon \sim \mathcal{N}(0,1)$.

The decoding module takes the $\mathbf{z}$ as input and then outputs the reconstructed adjacent matrix $\widehat{\mathbf{A}}$.
So the decoding step can be formulated as:
\begin{equation}
    \widehat{\mathbf{A}} = \sigma(\mathbf{z}\mathbf{z}^T)
\end{equation}
where $\sigma$ represents the decoding layer is activated by sigmoid function. 
Moreover, $\mathbf{z}\mathbf{z}^T$ can be converted to $\left \|\mathbf{z}\right \| \left \|\mathbf{z}^T\right\| \cos\theta$.
The inner product operation is beneficial capture the spatial correlation among different contexts.

During the training phase, we minimize the joint loss function $\mathcal{L}$ that is denoted as:
\begin{equation}
        \mathcal{L} = \sum \limits_{i=1}^{N} \underbrace{
        KL[q(\mathbf{z}|\mathbf{X},\mathbf{A}) || p(\mathbf{z})]
        }_{\text{KL Divergence between $q(.)$ and $p(.)$}}
        +
        \overbrace{
        \sum_{j=1}^{S} \left \| \mathbf{A}-\widehat{\mathbf{A}} \right \|^2 
        }^{\text{Loss between $\mathbf{A}$ and $\widehat{\mathbf{A}}$}}
    \label{equ:loss}
\end{equation}
where $N$ is the dimension of $\mathbf{z}$;
$S$ is the total number of the nodes in $\mathbf{A}$;
$q$ represents the real distribution of $\mathbf{z}$;
$p$ represents the prior distribution of $\mathbf{z}$.
$\mathcal{L}$ includes two parts, the first part is the Kullback-Leibler divergence between the standard prior distribution $\mathcal{N}(0,1)$ and the distribution of $\mathbf{z}$, and the second part is the squared error between $\mathbf{A}$ and $\widehat{\mathbf{A}}$.
The training process try to make the $\widehat{\mathbf{A}}$ close to $\mathbf{A}$ and let the the distribution of $\mathbf{z}$ similar to $\mathcal{N}(0,1)$.

Finally, we utilize global average aggregation for $\mathbf{z}$ to get the graph level representation, which is the latent representation of all context environments.

\subsection{Land-use Configuration Quantification and the Quality Measurement}

\begin{figure}[!thbp]
    \centering
    \vspace{-0.3cm}
    \includegraphics[width=0.9\linewidth]{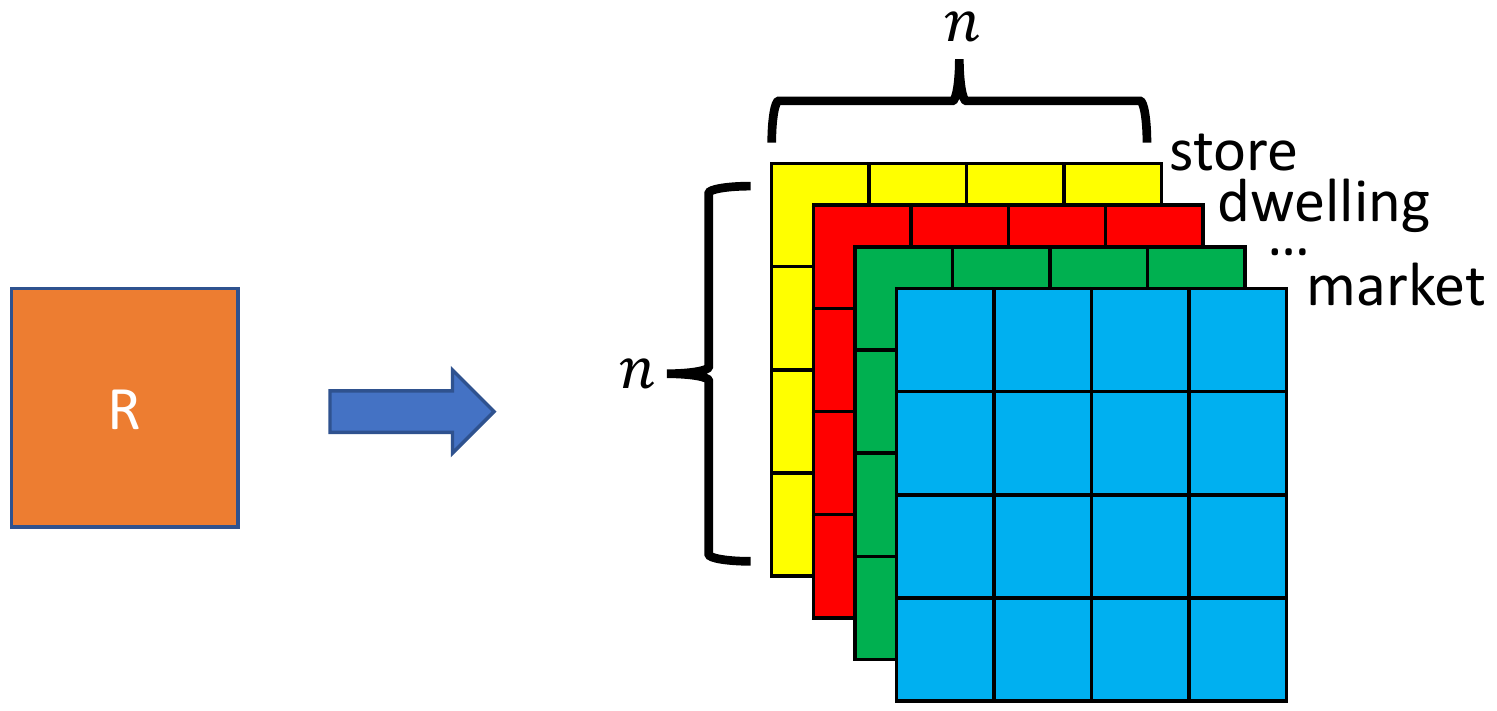}
    \vspace{-0.2cm}
    \caption{The construction of longitude, latitude, channel configuration tensor  where the value of each entry is the number of POIs with respect to a specific category in a specific latitude range and a specific longitude range.}
    \label{poi_dis}
    \vspace{-0.5cm}
\end{figure}

Land-use configuration indicates the location of different types of POIs, which expects an appropriate format of quantification for accommodating with a learning model.
To that end, we regard the POI distribution of one area as the land-use configuration
And then, we construct a multi-channel tensor to represent the land-use configuration, where each channel is the POI distribution across the geospatial area corresponding to one POI category. 
Figure \ref{poi_dis} shows an example of a land-use configuration. 
We first divide an unplanned area into $n\times n$ squares, then we sum up the number of each POI category in each square respectively.
Here, one POI category constructs one channel of the land-use configuration solution.
We obtain a land-use configuration as a multi-channel tensor.
  
Next, the other big challenge is how to evaluate the quality of land use configuration of the residential community? 
Because the urban planning is a complex field, urban planning specialists always evaluate the quality of land-use configuration solution from multiple aspects.
In our framework, we provide a quality hyper-parameter $Q$ for users, they can set the value of $Q$ to distinguish the quality of land-use configuration solution. 
In our experiment, we choose the POI diversity and the check-in frequency of an area as the quality standard.
Formally, we first calculate the total number of mobile check-in events of an area, denoted by $freq$, and the diversity of POI of an area, denoted by $div$.
We then incorporate the two indicators into together by $Q=\frac{2\times freq \times div}{freq+div}$ \cite{wang2018ensemble}.
If $Q > 0.5$, the solution is regarded as an excellent solution. 
Otherwise, it is justified as a terrible solution.

\subsection{Generating Excellent Land-use Configuration Solution by GAN}

Generative adversarial networks (GAN) is a popular deep generative model.
The framework of this technique is suitable to generate realistic data samples via an adversarial way. 
In the computer vision field, GAN achieves tremendous achievements.
So, here, we utilize the GAN framework to generate excellent land-use configuration solutions of an unplanned area according to the representation of the context environments.

\begin{figure*}[t]
\vspace{-0.3cm}
    \centering
    \includegraphics[width=0.75\linewidth]{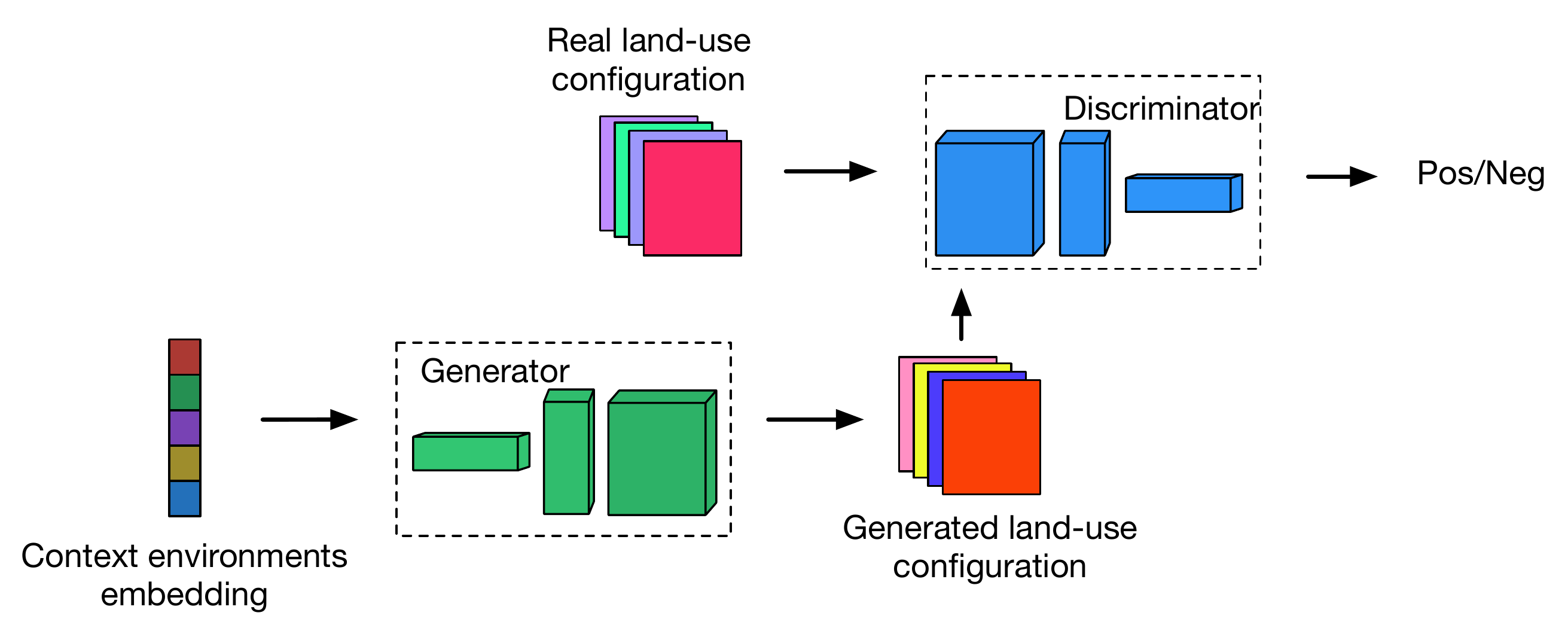}
    \caption{Automatic land-use configuration planner}
    \vspace{-0.5cm}
    \label{GAN_framework}
\end{figure*}

Figure \ref{GAN_framework} represents the structure of our automated land-use configuration planner.
In common, the real land-use configuration includes two categories: excellent and terrible.
The purpose of the automated planner is to generate the excellent land-use configuration plan based on the context embedding.
Formally, we input the context embedding into the generator to generate the land-use configuration solution.
In order to improve the generative ability, the discriminator classify the excellent plans as positive and the terrible plans as negative.
Algorithm \ref{alg:auto_planner} shows the detail information about the training phase.

\begin{algorithm}[hbtp] 
  \tcp{start training.}
  \For{number of training iterations}{
  \tcp{update discriminator firstly.}
  \For{n steps}{
    Sample minibatch of $m$ excellent land-use configuration samples $\left \{ \mathbf{E}^1,\mathbf{E}^2,...,\mathbf{E}^m \right \}$.\\
    Sample minibatch of $m$ context information embedding samples $\left \{ \mathbf{z}^1,\mathbf{z}^2,...,\mathbf{z}^m \right \}$. \\
    Generate land-use configuration samples by generator,  $\left \{ \mathbf{F}^1,\mathbf{F}^2,...,\mathbf{F}^m \right \}$. Here, $\mathbf{F}^i=G(z^i)$.\\
    Sample minibatch of $m$ terrible land-use configuration samples $\left \{ \mathbf{T}^1,\mathbf{T}^2,...,\mathbf{T}^m \right \}$. \\
    Update the discriminator by ascending its gradient:\\
    ~\\
    $\bigtriangledown_{\theta_d} \frac{1}{m}\sum_{i=1}^{m} [ log(D(\mathbf{E}^i)) + log(D(1-\mathbf{F}^i))$\\
    $+ log(D(1-\mathbf{T}^i)) ].$
    ~\\
  }
  \tcp{update generator secondly.}
    Sample minibatch of $m$ context information embedding samples
    $\left \{ \mathbf{z}^1,\mathbf{z}^2,...,\mathbf{z}^m \right \}$. \\
    Update the generator by descending its gradient:\\
    ~\\
    $\bigtriangledown_{\theta_g} \frac{1}{m} \sum_{i=1}^{m} log(1-D(G(\mathbf{z}^i))).$
    ~\\
  }
 
  \caption{Minibatch adaptive moment estimation training of automatic land-use configuration model. We adjust one hyperparameter $f$ to change the update frequencies of the weight of the discriminator.}
  \label{alg:auto_planner}
\end{algorithm}

In algorithm \ref{alg:auto_planner}, we first update the parameters of the discriminator module and fix the parameters of the generator module. 
We then feed the excellent, terrible, and generated land-use configuration samples into the discriminator module.
Next, the discriminator outputs the classification result that is activated by the sigmoid function.
It gives higher classification scores for excellent samples than terrible and generated samples.
Next, we fix the discriminator module and update the parameter of the generator module.
The contexts embedding vectors are feed into the generator, and output generated land-use configuration solutions.
Afterward, the generated solutions are feed into the discriminator to justify the quality of them.
We update the parameter of the generator module to improve the generated ability of itself.
The update gradient comes from the justification result of the discriminator module. 
Finally, We obtain one automatic land-use configuration planner when the GAN model converges. 
If we obtain the context embedding, the discriminator can generate one excellent land-use configuration for the unplanned area.

\section{Experiment Results}

In this section, we conduct extensive experiments and case studies to answer the following questions:
\begin{enumerate}
    \item Does our proposed automatic planner (LUCGAN) outperform the baseline methods?
    \item What is the difference between the context of excellent land-use configuration plans and terrible plans?
    \item How does generated land-use configuration plans look like?
    \item How many proportions of each POI category occupies in generated plans?
    \item What is the generated situations of different categories in the generated plan?

\end{enumerate}

\subsection{Data Description}

We use the following datasets for our study:

\begin{enumerate}
    \item \textbf{Residential Community:} 
    The residential community dataset contains 2990 residential communities in Beijing, where each residential community is associated with the information of its latitude and longitude. 
    \item \textbf{POI:} The Beijing POI dataset includes 328668 POI records from 2011, where each POI item includes its latitude, longitude and the category.
    The POI information is shown in Table \ref{poi_lists}.
    
    \item \textbf{Taxi Trajectories:}
    The taxi trajectories dataset are collected from a Beijing taxi company, where each record contains trip ID, distance(m), travel time(s), average speed(km/h), pick-up and drop off time, pick-up and drop-off point.

    \item \textbf{Public Transportation:}
    This dataset logs the transactions of buses in Beijing between 2012 and 2013. After analyzing the dataset, it contains 1734247 bus trips, 718 bus lines.
    We use this dataset to obtain the public transportation situation.

    \item \textbf{House Price:}
    The house price dataset includes continuous five months house price data of each residential community in Beijing between 2011 and 2012, which is collected from Soufang website.
    
    \item \textbf{Check-In:}
    This dataset is the weibo check-in records in Beijing between 2011 and 2013, where each check-in item has its longitude, latitude, check-in time and check-in place.
    We utilize this dataset to analyze the vibrancy of one area.
    
\end{enumerate}

\begin{table}[htbp]
\vspace{-0.4cm}
\small
\centering
\setlength{\abovecaptionskip}{0.cm}
\caption{POI category list}
\setlength{\tabcolsep}{1mm}{
\begin{tabular}{cccccc}  
\toprule
 code  & POI category & code & POI category  \\  
\midrule       
  0  & road & 10 & tourist attraction \\
  1 & car service & 11 & real estate \\
  2 & car repair & 12 & government place \\
  3 & motorbike service & 13 & education \\
  4 & food service & 14 & transportation \\
  5 & shopping & 15 & finance\\
  6 & daily life service & 16 & company\\
  7 & recreation service & 17 & road furniture\\
  8 & medical service & 18 & specific address \\
  9 & lodging  & 19 & public service\\
\bottomrule
\end{tabular}}
\label{poi_lists}
\vspace{-0.4cm}
\end{table}

\subsection{Evaluation Metrics}

Because evaluating the quality of the urban land-use configuration is an open question, there is no standard measurement. 
In this paper, we evaluate the quality of generated planning solution from multiple aspects to express the effectiveness of our framework:
(1)\textbf{Scoring Model}.
We build a random forest model based on the excellent and terrible land-use configuration plans.
The model is capable of giving higher scores for excellent land-use configuration plans and provide lower scores for terrible plans.
When we get the generate land-use configuration solutions, the scoring model can be utilized to quantify the quality of the generated solutions.
(2) \textbf{Visualization}. 
In order to explore the generated solutions, we select one representative sample to visualize from multiple aspects.
We can observe the solutions directly in this way.
It is helpful to learn the difference between our planner and other baselines .

\subsection{Baseline Methods}
We compare the performances of our framework(LUCGAN) against the following three baseline methods:
\begin{enumerate}
    \item \textbf{VAE}: is an encoder-decoder paradigm algorithm.
    The encoder encodes image data into latent embedding; the decoder decodes the embedding into the original data.
    In this experiment, we input excellent land-use configuration into VAE to learn the distribution of excellent solutions by minimizing the reconstruction loss.
    Then, we utilize the decoder to generate the solution based on the context environment embedding when VAE converges. 
    
    \item \textbf{AVG}: 
    generates the land-use configuration by calculating the mean value of all excellent land-use plans, which reflects the average level of all excellent samples.
    But this method can not provide a customized solution based on different context environments.
    
    \begin{figure}[htbp]
    \centering
    \includegraphics[width=0.8\linewidth]{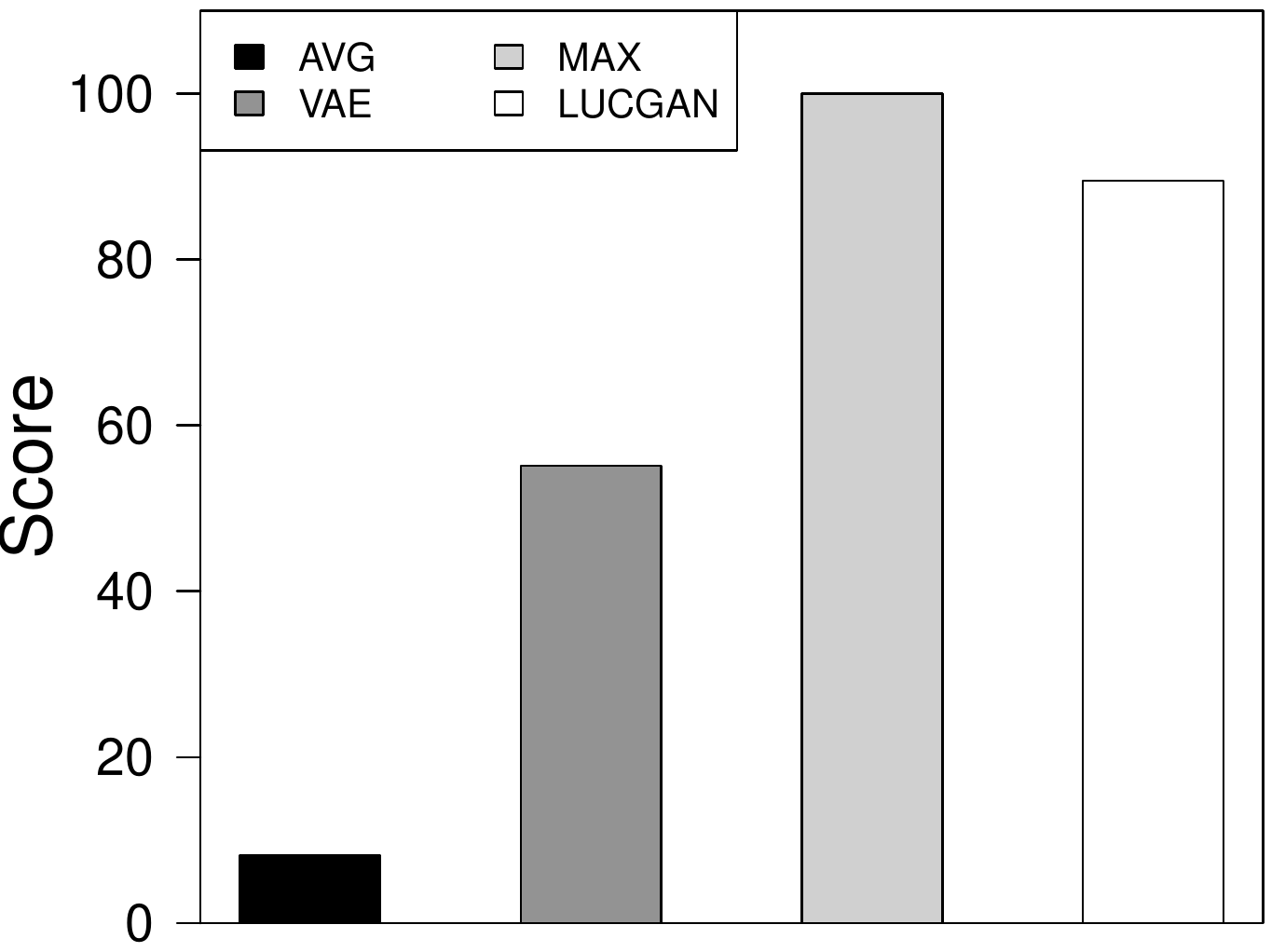}
    \vspace{-0.3cm}
    \caption{The quality score for different generated methods.}
    \label{score_model}
    \vspace{-0.8cm}
    \end{figure}
    
    \begin{figure}[htbp]
    \centering
    \includegraphics[width=0.8\linewidth]{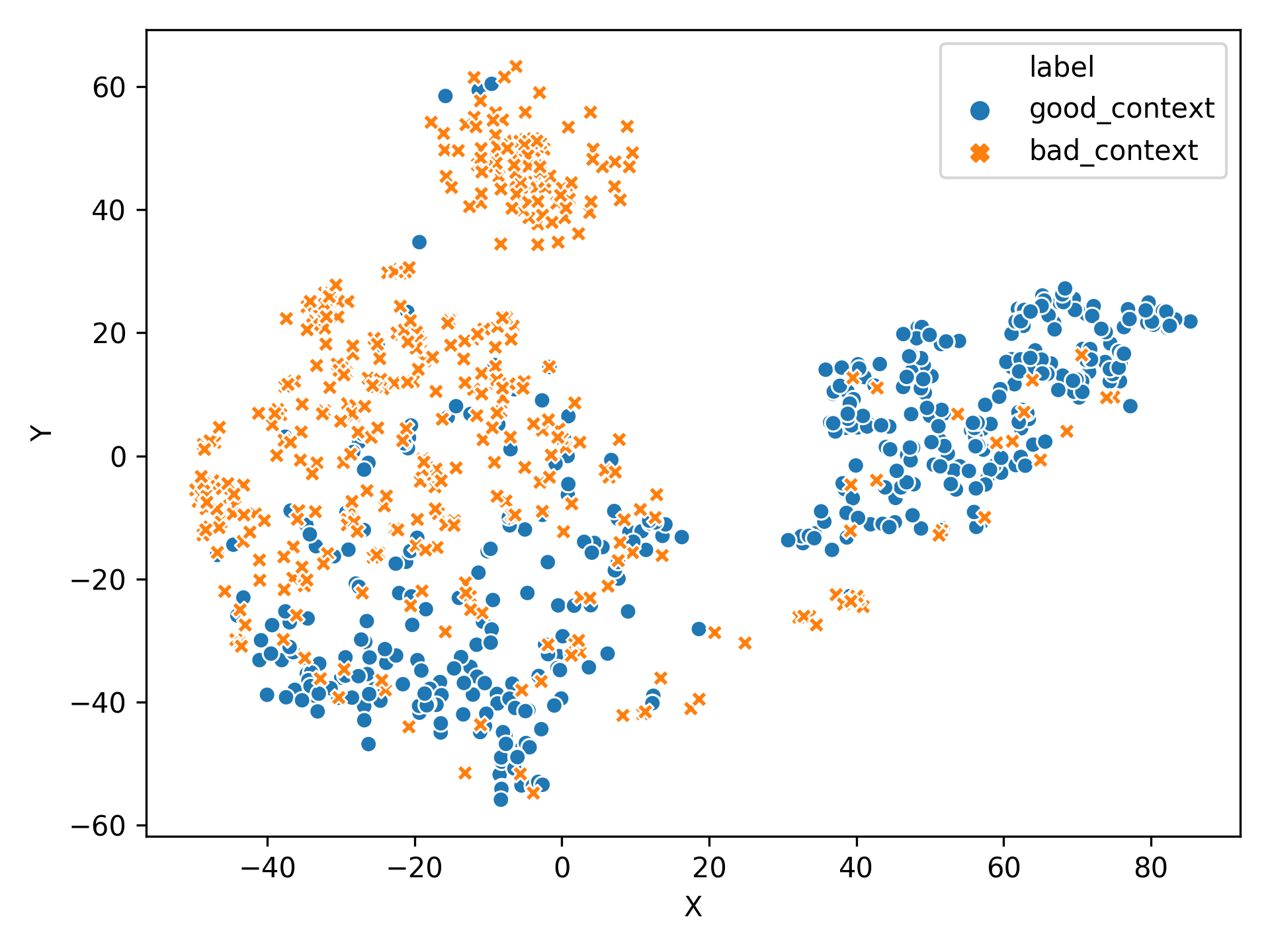}
    \vspace{-0.6cm}
    \caption{Visualization for different contexts.}
    \label{emb_context}
    \vspace{-0.55cm}
    \end{figure}

    \item \textbf{MAX}: 
    generates the land-use configuration solutions by applying max operation on all excellent land-use plans.
    The result of this method reflects the most dominated POI categories in each geographical block.
    The same to AVG, MAX also can not generate a customized solution based on different context environments. 
    
\end{enumerate}

We conduct all experiments on a x64 machine with Intel i9-9920X 3.50GHz CPU, 128GB RAM and Ubuntu 18.04.

\begin{figure*}[!thb]
\setlength{\abovecaptionskip}{-2pt} 
	\centering
	\subfigure[LUCGAN]{\label{fig:LUCGAN}\includegraphics[width=0.33\linewidth]{{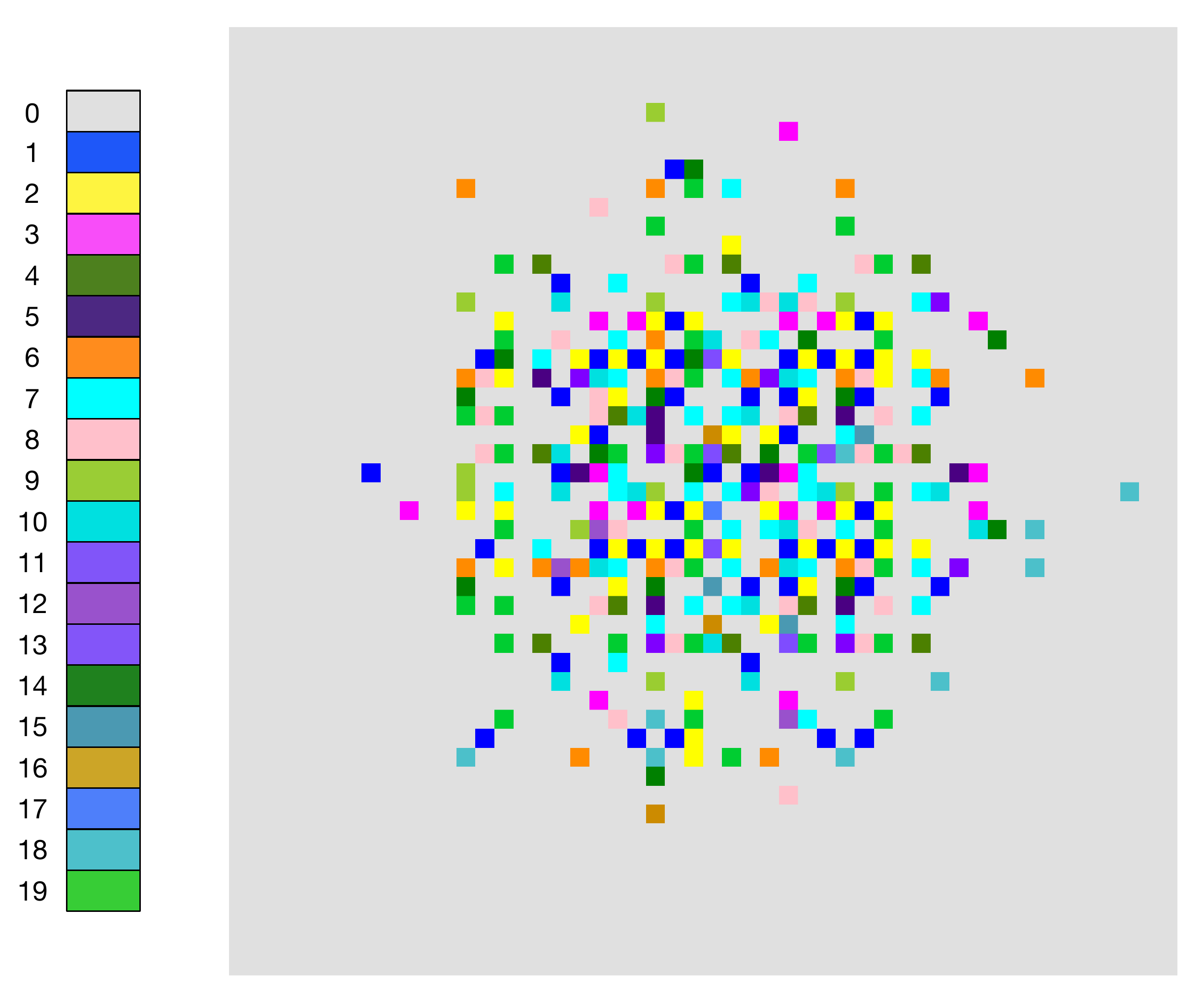}}}
	\subfigure[VAE]{\label{fig:vae}\includegraphics[width=0.33\linewidth]{{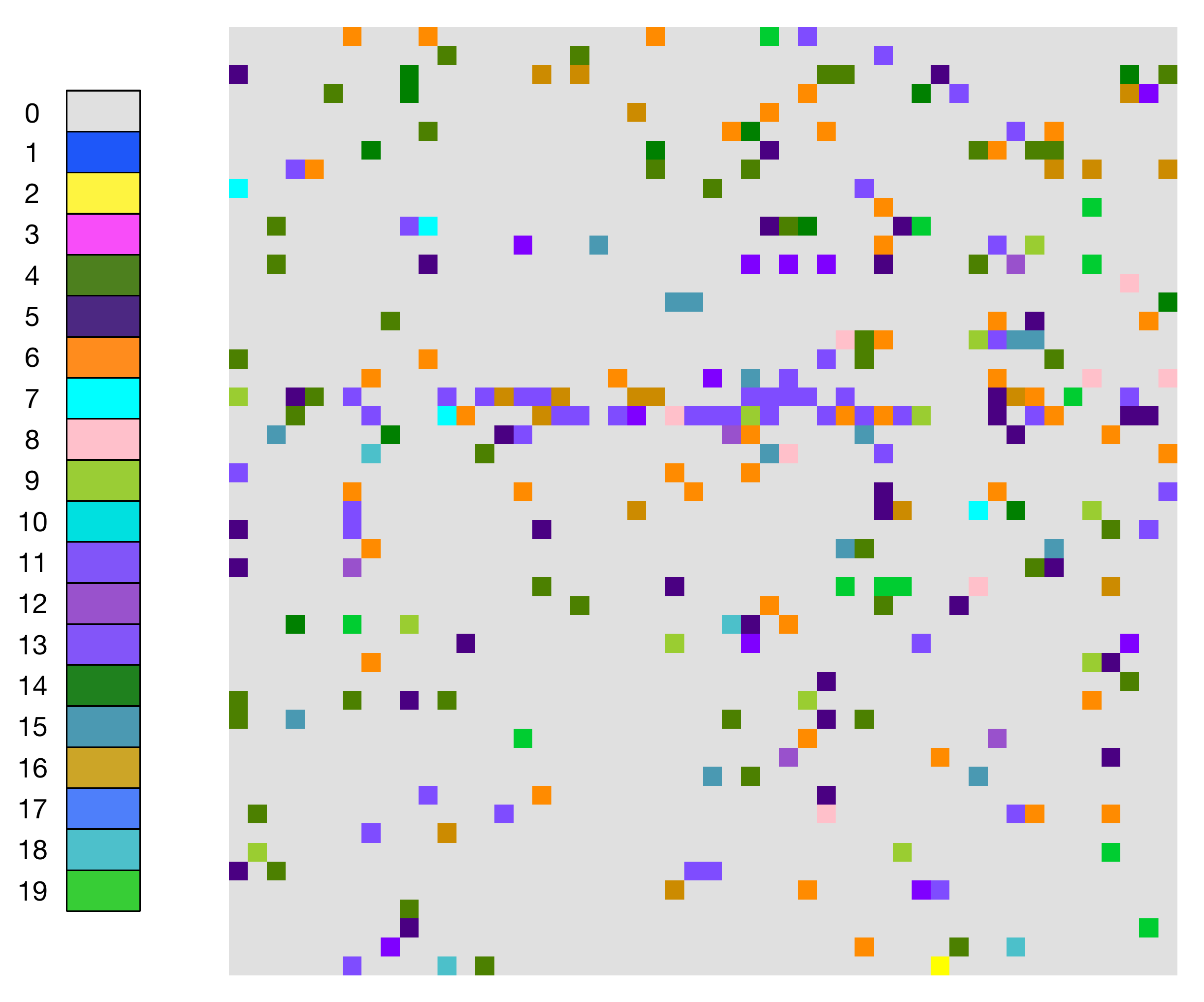}}}
	\subfigure[MAX]{\label{fig:max}\includegraphics[width=0.33\linewidth]{{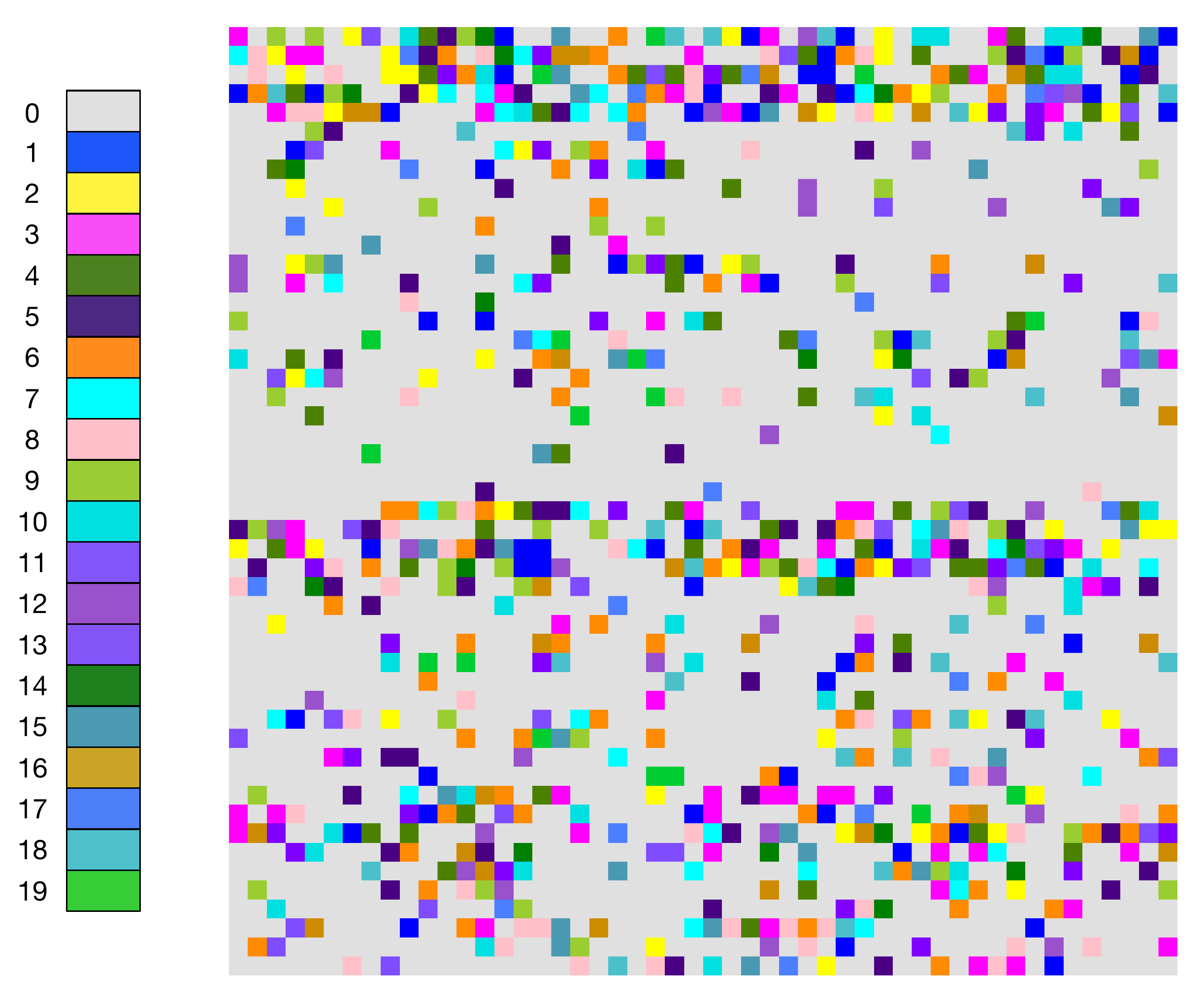}}}
	\caption{Comparison of different generated land-use configuration solutions by different generated methods.}
	\label{fig:generated_solution}
	\vspace{-0.5cm}
\end{figure*}

\begin{figure*}[!thb]
\setlength{\abovecaptionskip}{-2pt} 
	\centering
	\subfigure[LUCGAN]{\label{fig:LUCGAN_poi}\includegraphics[width=0.33\linewidth]{{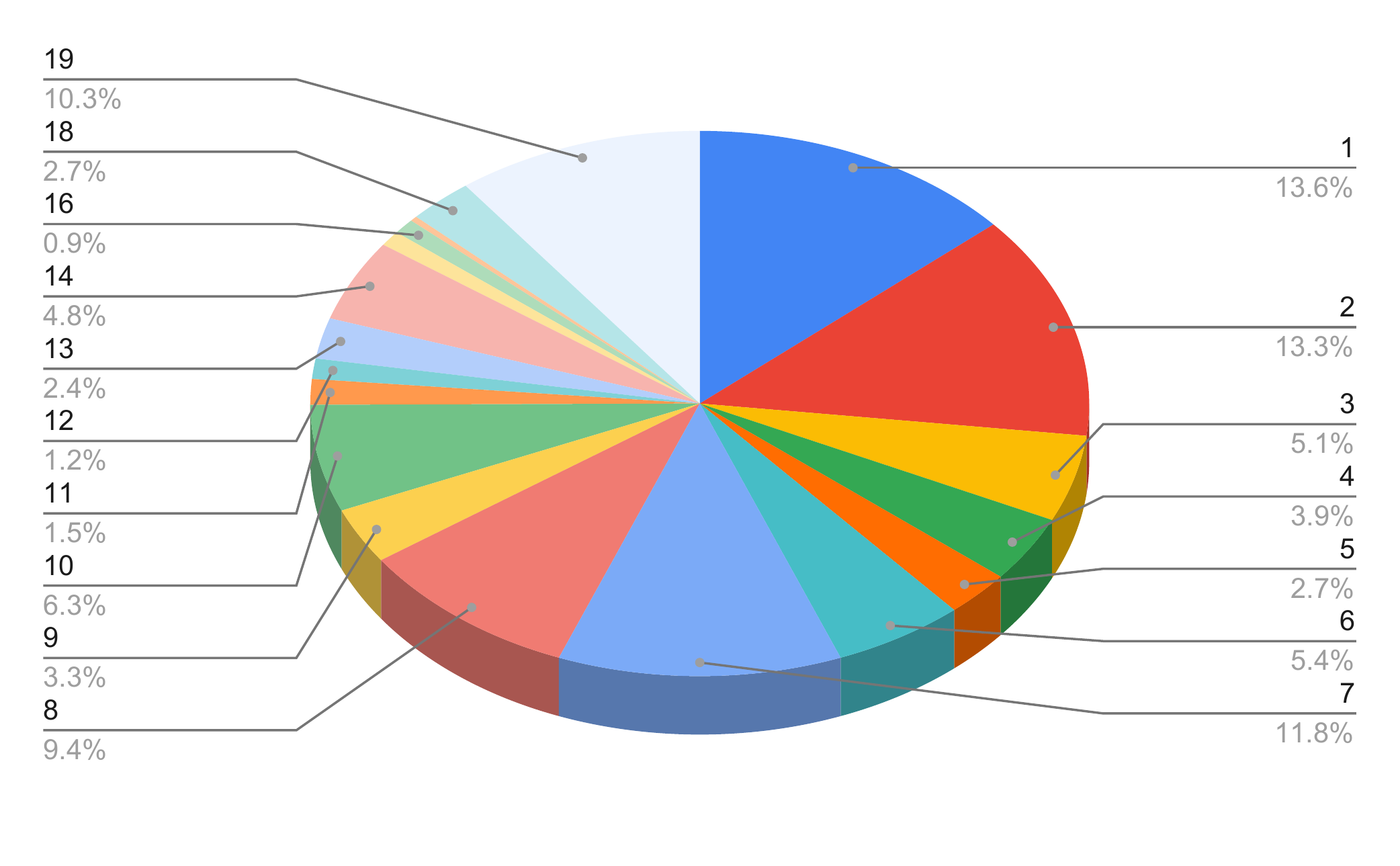}}}
	\subfigure[VAE]{\label{fig:vae_poi}\includegraphics[width=0.33\linewidth]{{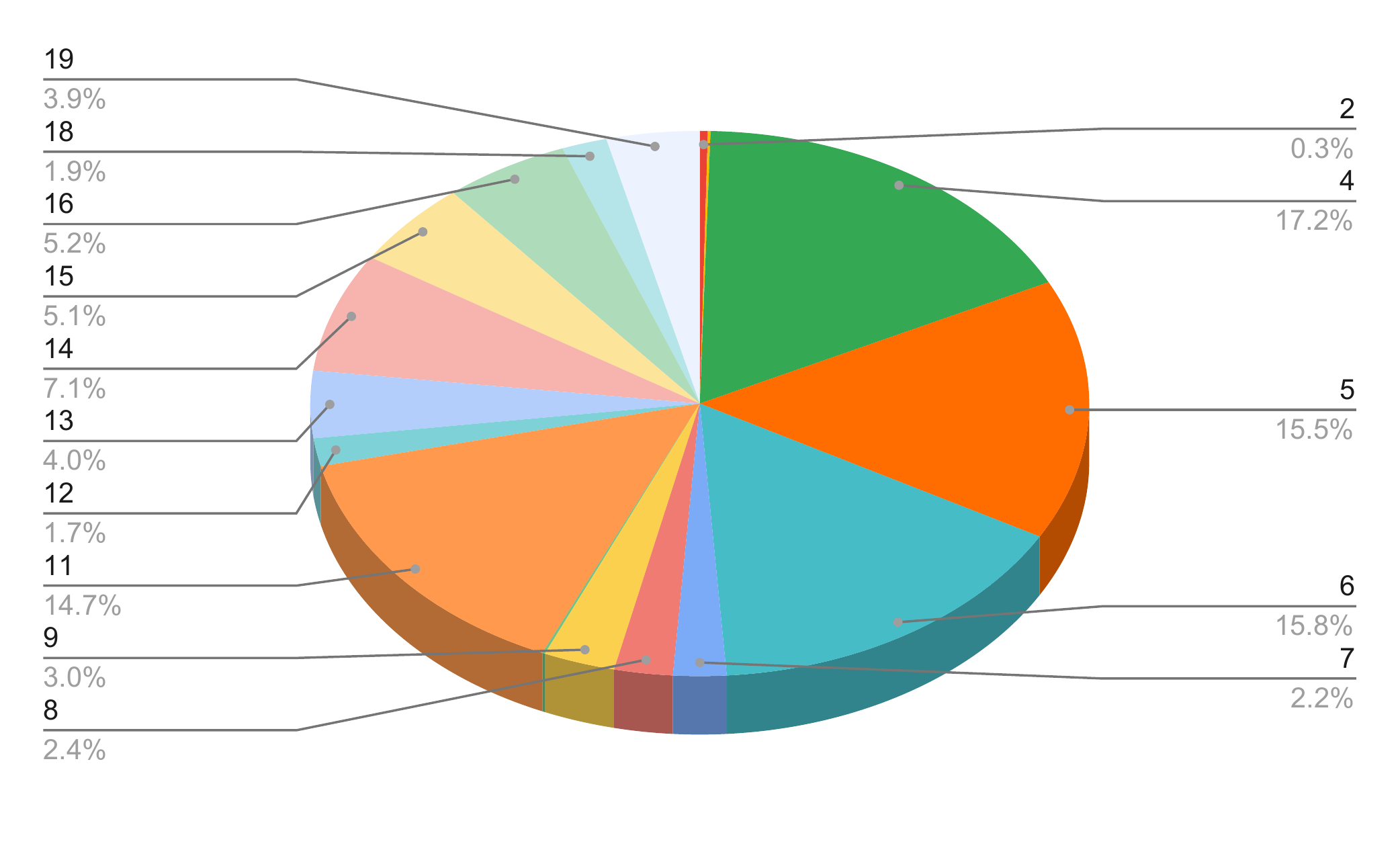}}}
	\subfigure[MAX]{\label{fig:max_poi}\includegraphics[width=0.33\linewidth]{{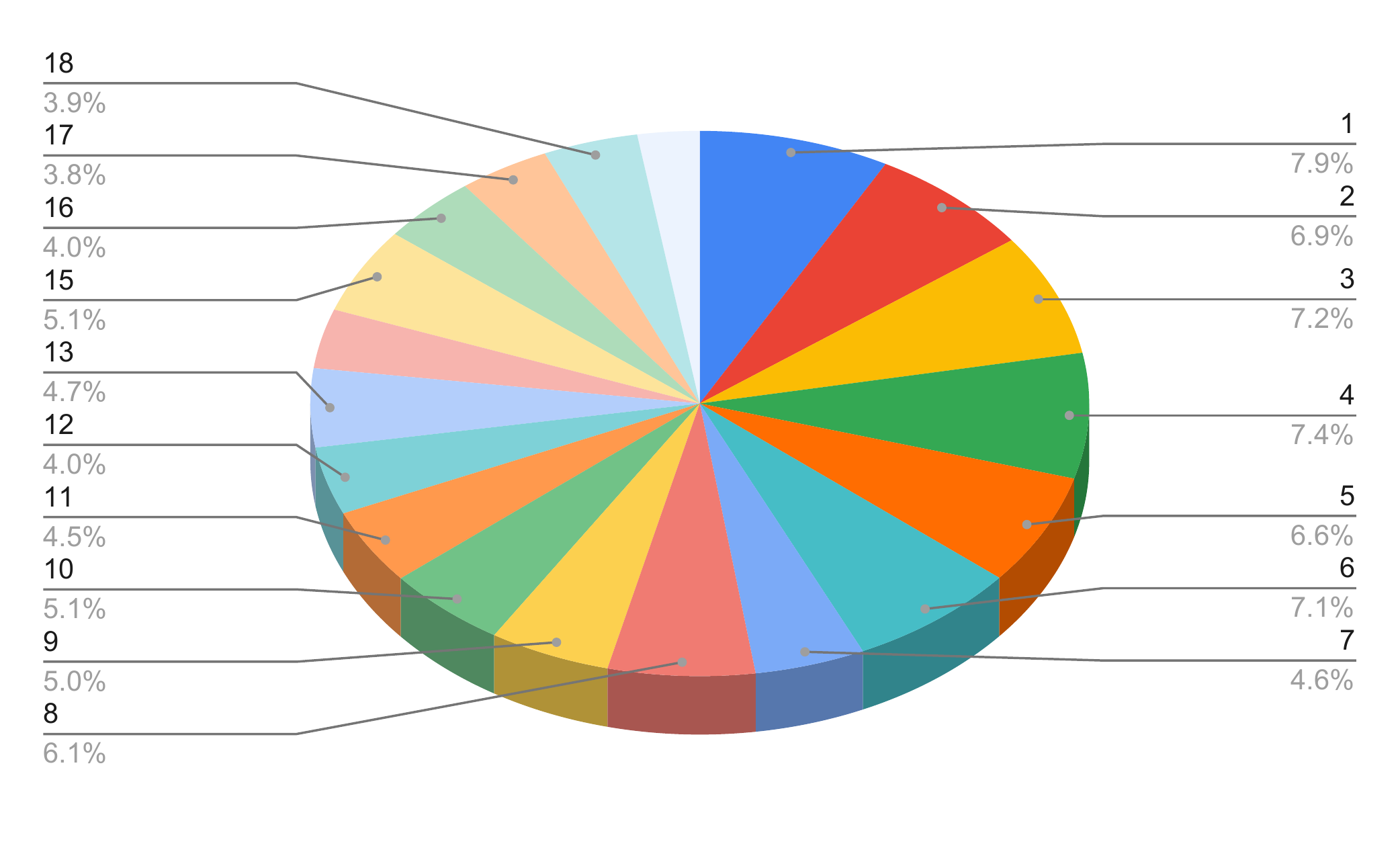}}}
	\caption{Comparison of the proportion of each POI category in different generated solutions.}
	\label{fig:category_result}
	\vspace{-0.5cm}
\end{figure*}

\subsection{Overall Performance}

Figure \ref{score_model} shows the quality score produced by the scoring model for different generated methods.
An interesting phenomenon is that the MAX method ranks the 1st compared with other methods. 
A possible explanation is the scoring model only captures the distribution of original excellent plans.
The MAX method incorporates all excellent plans by max operation, so the generated solution reflects the dominated POI categories of each geographical block, which is also inherent in the original data distribution.
Therefore, the scoring model gives the highest score for the MAX method.
Although the MAX method ranks the 1st, it does not indicate the MAX method is better than LUCGAN.
This is because the MAX method only produces one kind of planning solution no matter what the context environment is.
But the LUCGAN can customize the solutions based on different context environment embedding.
In addition, the score of LUCGAN is also high, which indicates the LUCGAN capture the intrinsic rule of excellent plan distribution.
So the LUCGAN is an effective and flexible generated method for generating land-use configuration.

\subsection{Study of the Context Environment}

Intuitively, the context environment is vital for generating the land-use configuration.
Different context environments produce different kinds of land-use configuration solutions.
In order to explore the difference between the context environments of excellent plans and bad plans, we utilize T-SNE algorithm on the context environment embedding.
We randomly choose 500 good and bad context embedding respectively to visualize.

Figure \ref{emb_context} is the visualization result of different kinds of context environments.
We find that the pattern of good contexts and the pattern of bad contexts are discriminative, which proves that it is reasonable to exploit the context environment for generating the land-use configuration solutions.

\subsection{Study of the Geographical Distribution Generated by Different Approaches}

In order to observe the generated land-use configuration solution clearly, we pick up a representative solution to visualize.
Owing to the generated solution has multiple channels and each channel owns lots of blocks, we merge these channels of the solution into one by we setting the dominated POI category as the final result for each geographical block.
The merged solution reflects POI distribution in geographical spatial space. 

Figure \ref{fig:generated_solution} shows the visualization result
Different color blocks represent different POI categories.
The generated solution by LUCGAN is regular and organized, different POI categories intersect with each other.
But the distribution of generated solutions by VAE and MAX are so chaotic that there no clear promising patterns.
This experiment indicates the better effectiveness of LUCGAN compared against other baseline methods.

\begin{figure*}[!thb]
\setlength{\abovecaptionskip}{-0pt} 
	\centering
	\subfigure[car service]{\label{fig:channel1}\includegraphics[width=0.24\linewidth]{{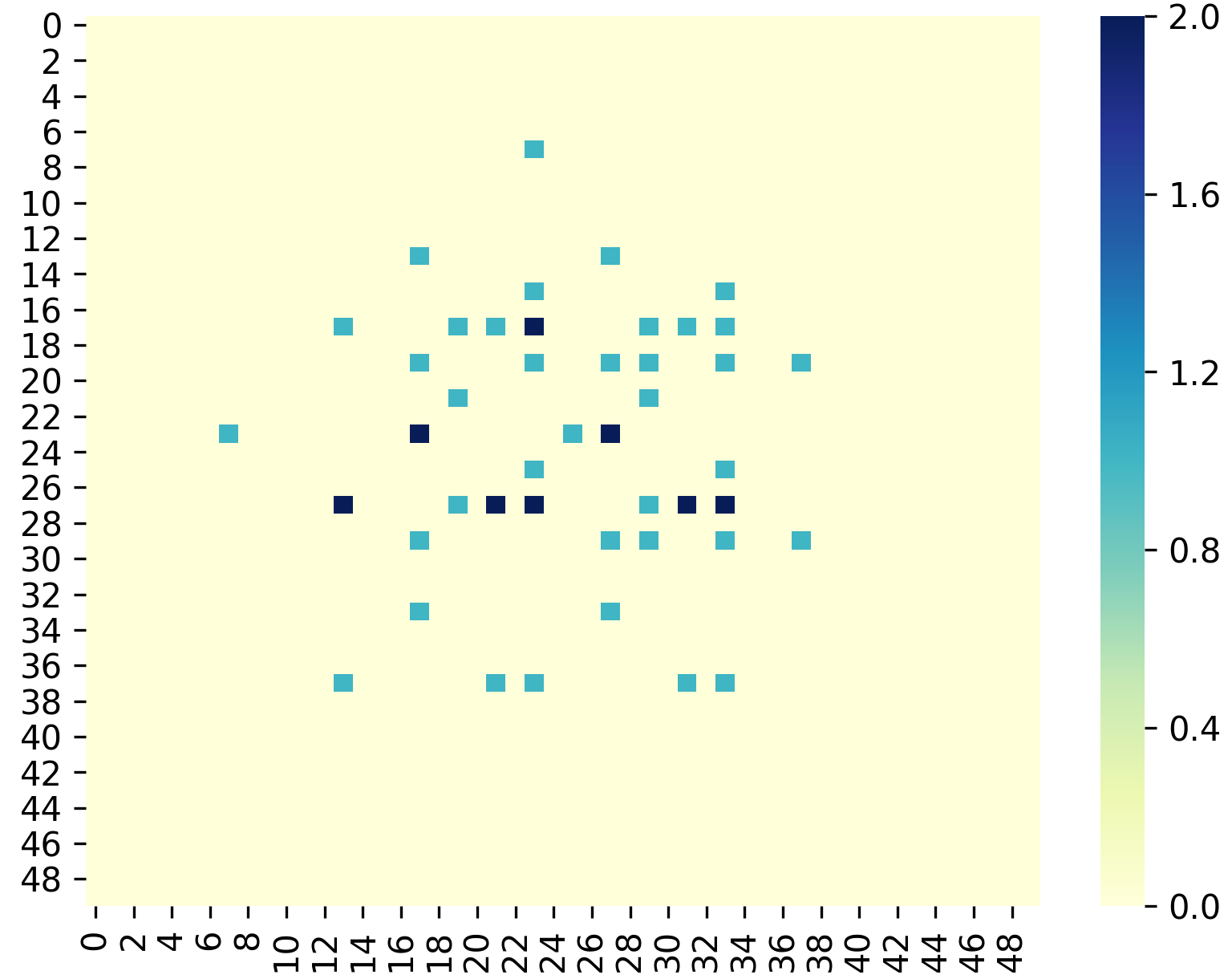}}}
	\subfigure[food service]{\label{fig:channel4}\includegraphics[width=0.24\linewidth]{{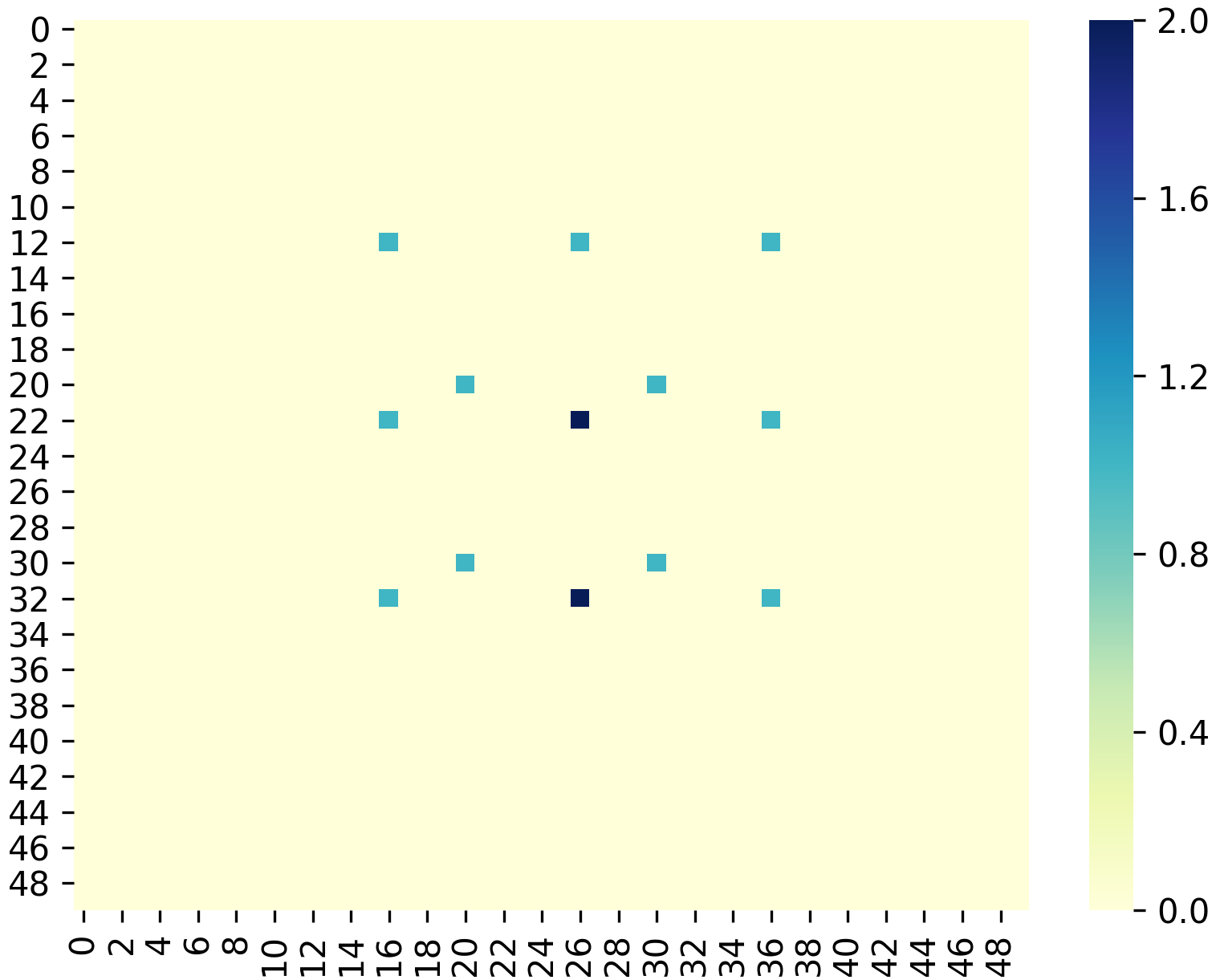}}}
	\subfigure[daily life service]{\label{fig:channel6}\includegraphics[width=0.24\linewidth]{{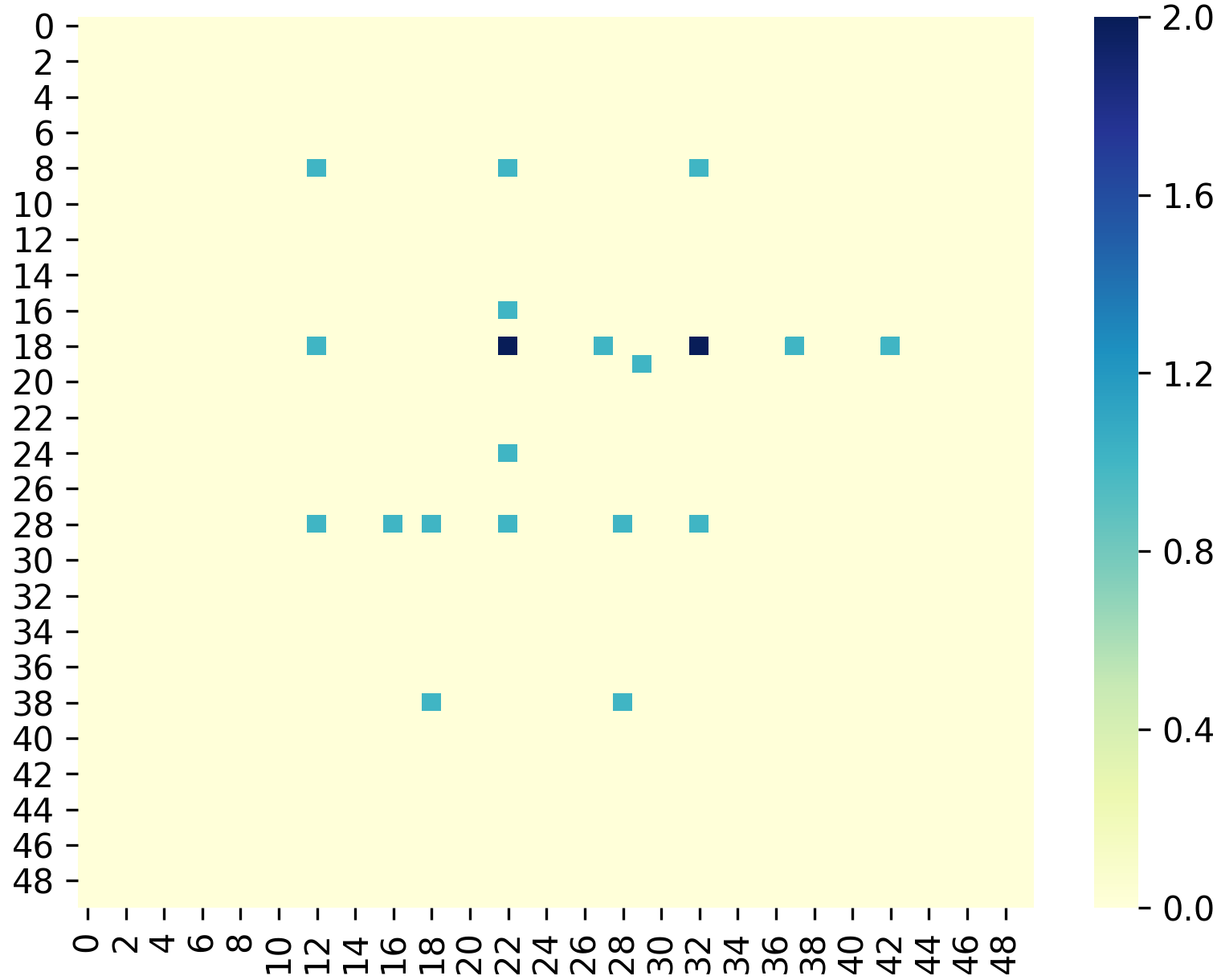}}}
	\subfigure[recreation service]{\label{fig:channel7}\includegraphics[width=0.24\linewidth]{{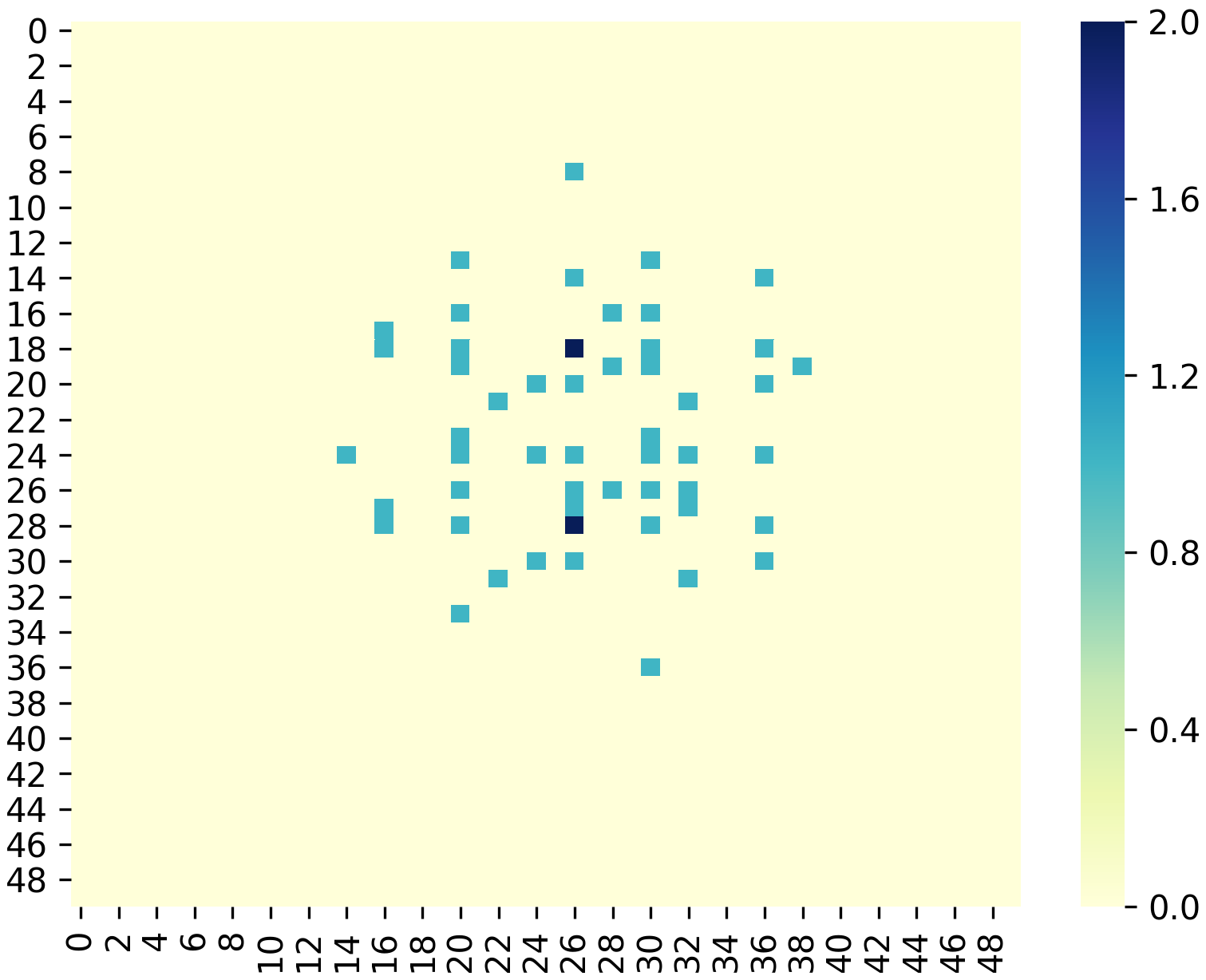}}}
	
	\vspace{-0.3cm}
	
	\subfigure[tourist attraction]{\label{fig:channel10}\includegraphics[width=0.24\linewidth]{{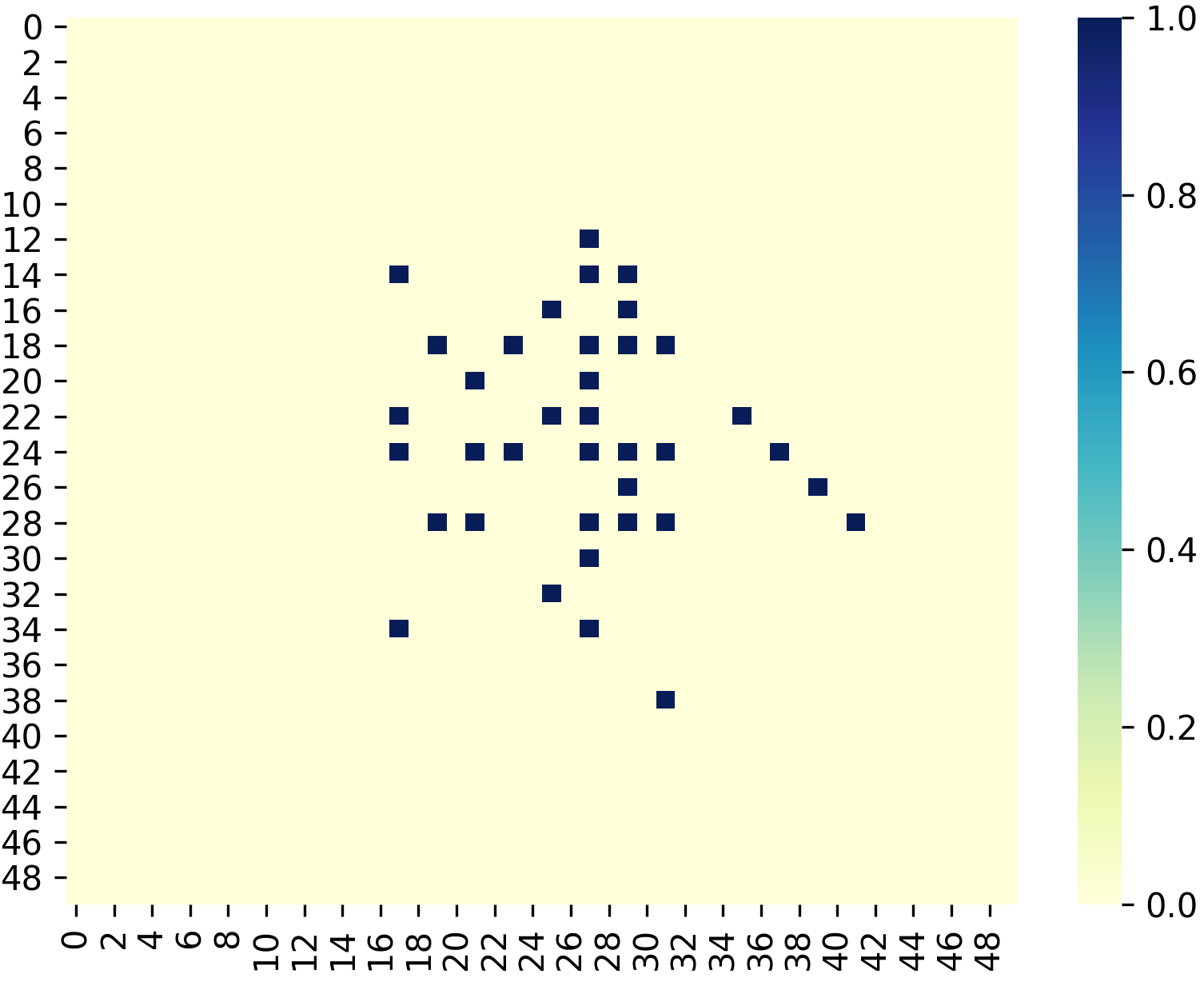}}}
	\subfigure[real estate]{\label{fig:channel11}\includegraphics[width=0.24\linewidth]{{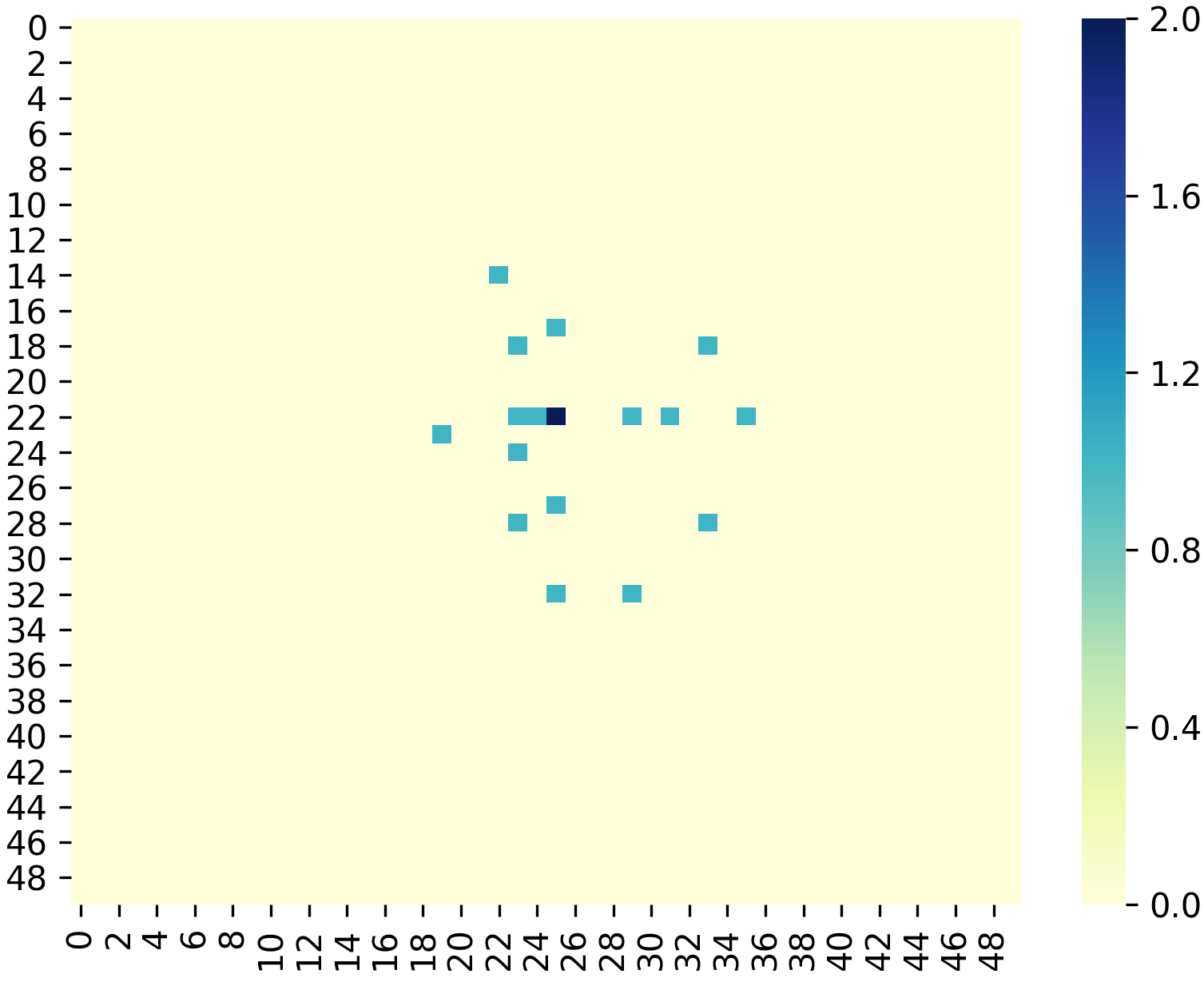}}}
	\subfigure[government place]{\label{fig:channel12}\includegraphics[width=0.24\linewidth]{{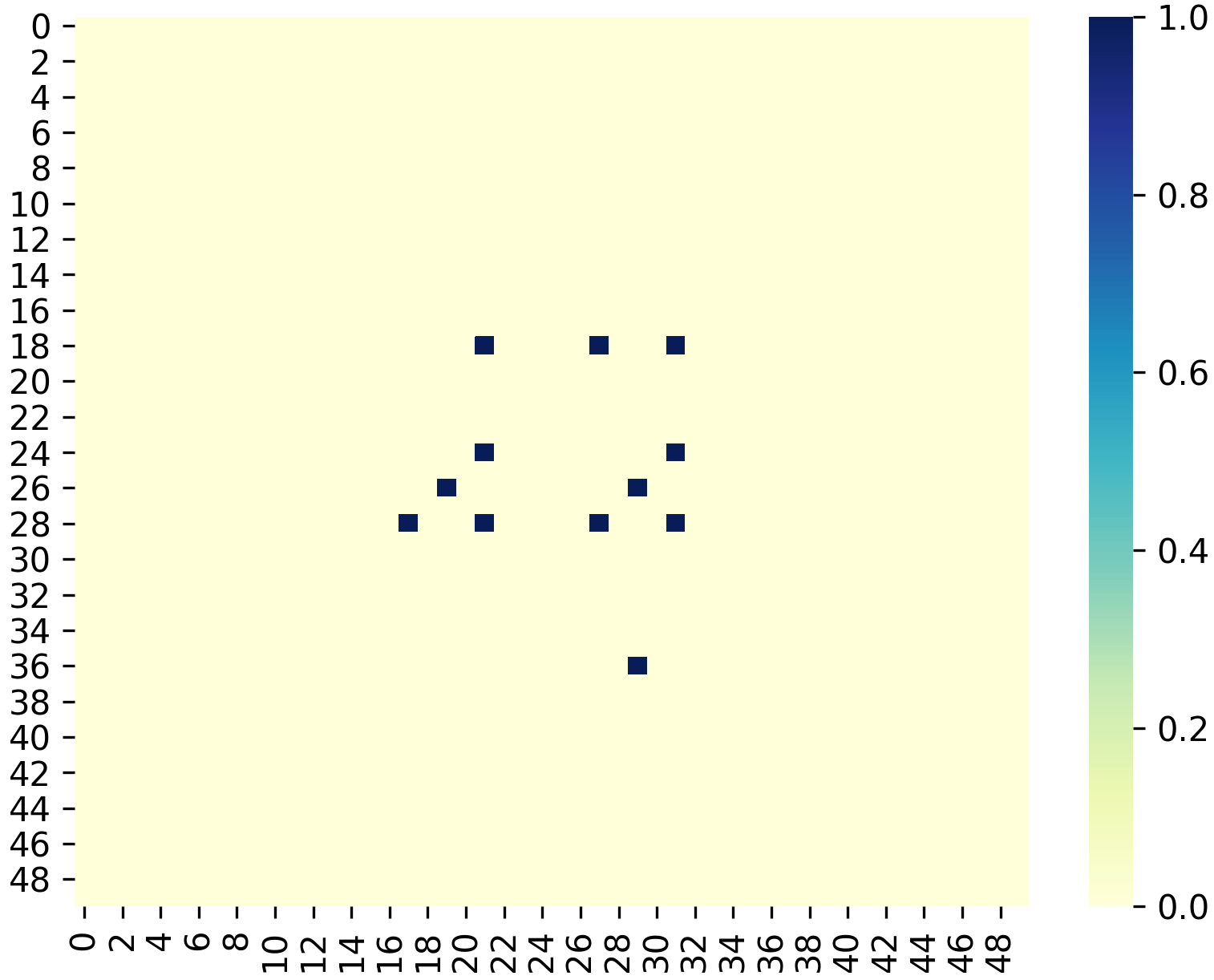}}}
	\subfigure[education]{\label{fig:channel13}\includegraphics[width=0.24\linewidth]{{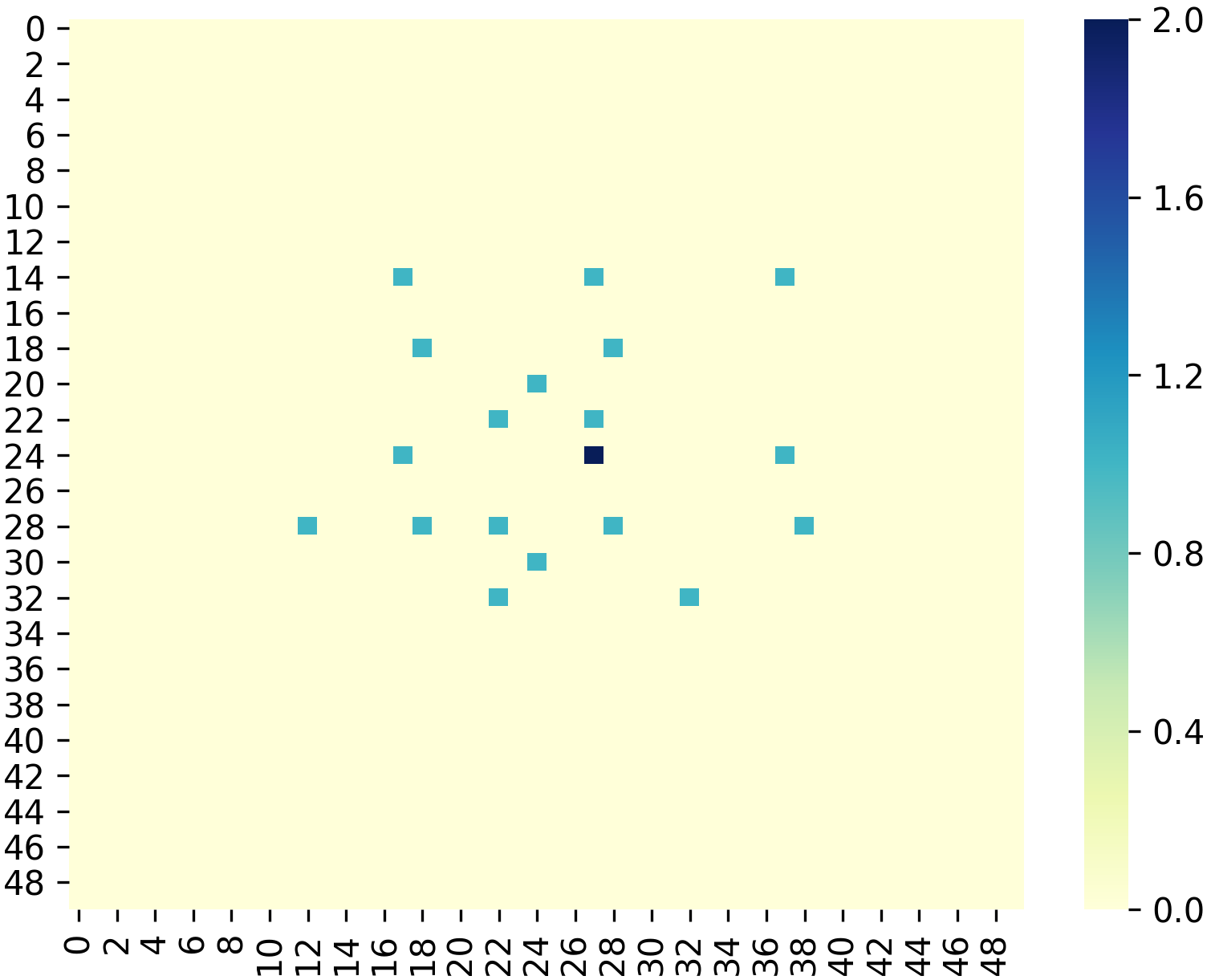}}}
	
	\vspace{-0.3cm}
	
	\subfigure[transportation]{\label{fig:channel14}\includegraphics[width=0.24\linewidth]{{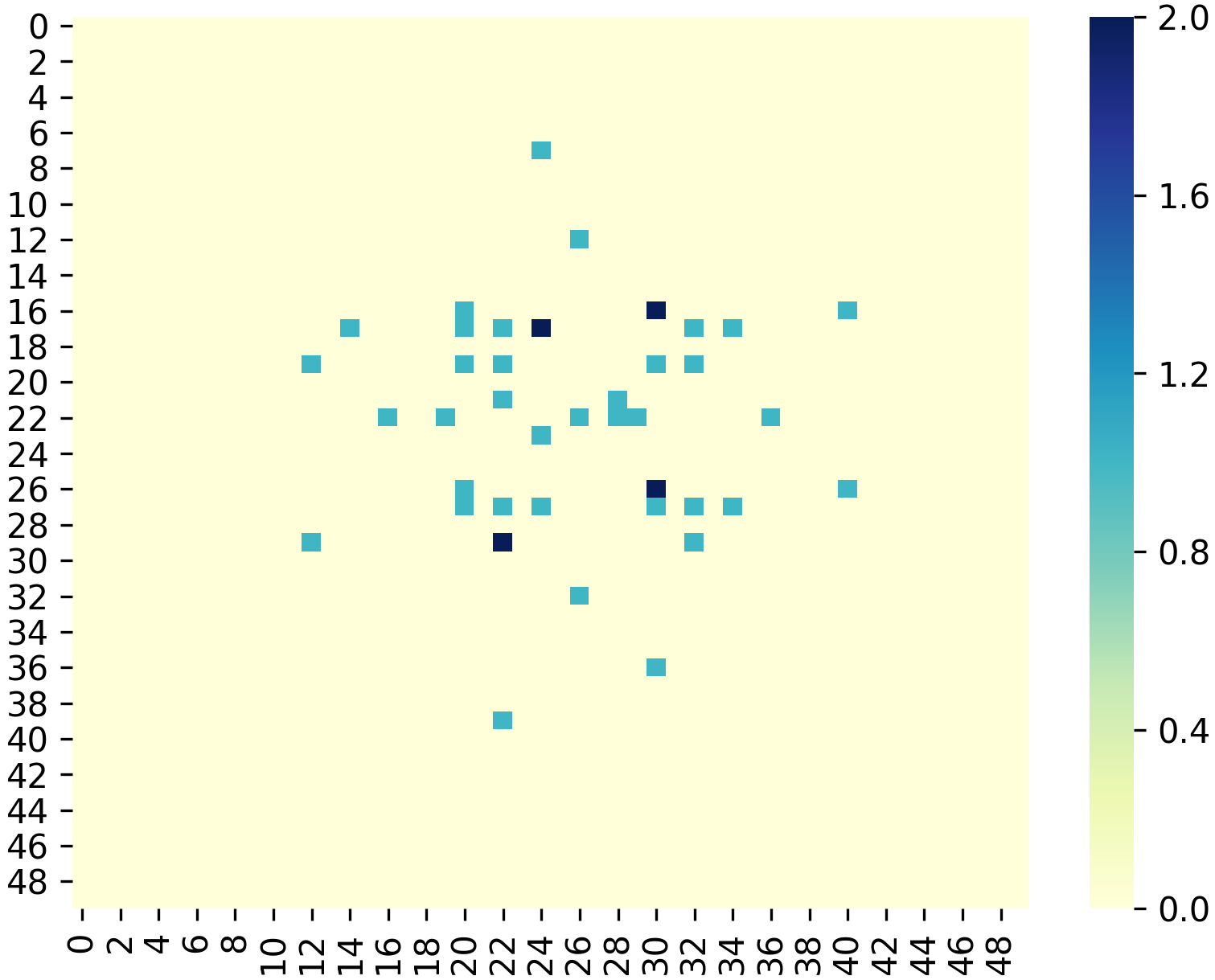}}}
	\subfigure[specific address]{\label{fig:channel18}\includegraphics[width=0.24\linewidth]{{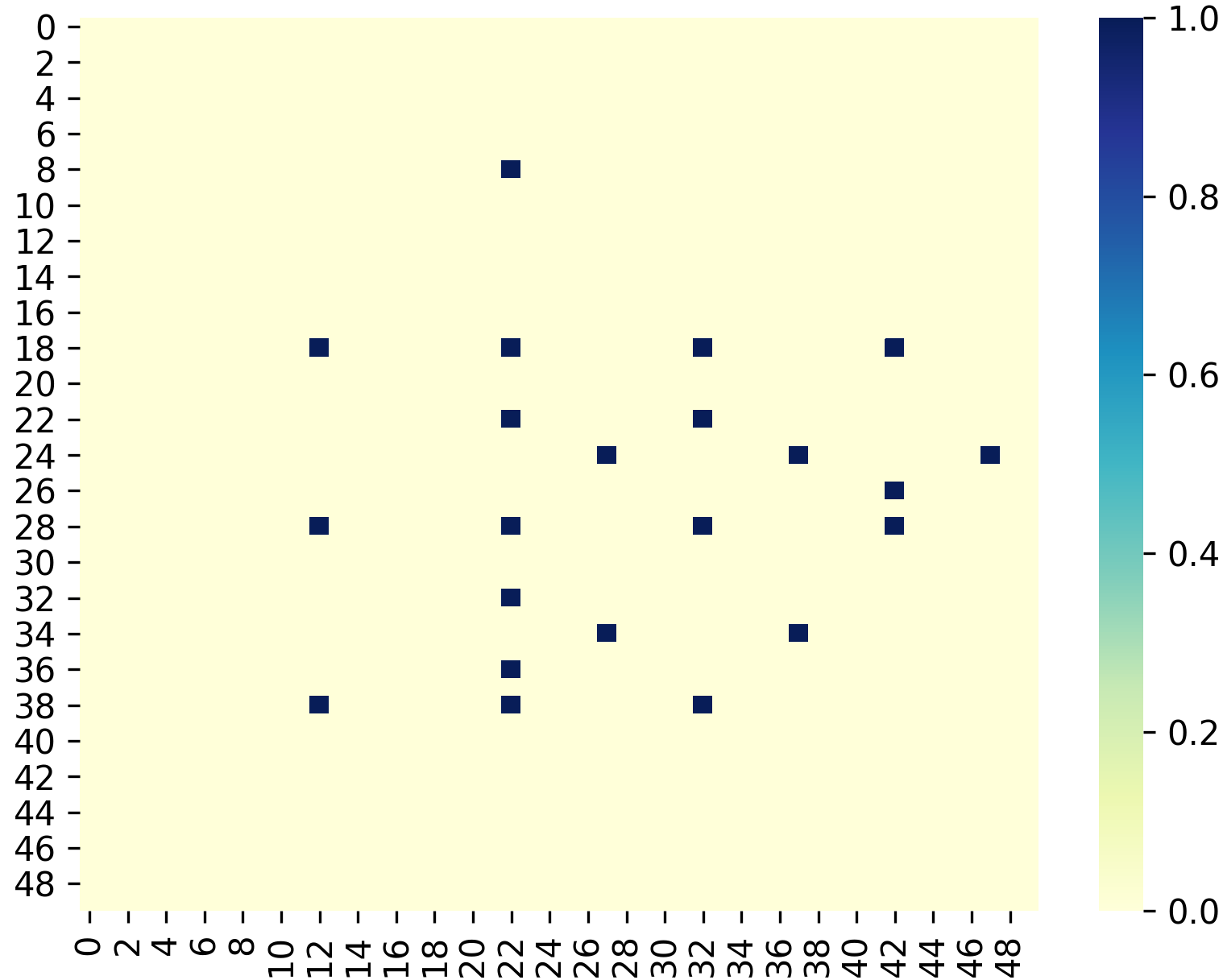}}}
	\subfigure[finance]{\label{fig:channel15}\includegraphics[width=0.24\linewidth]{{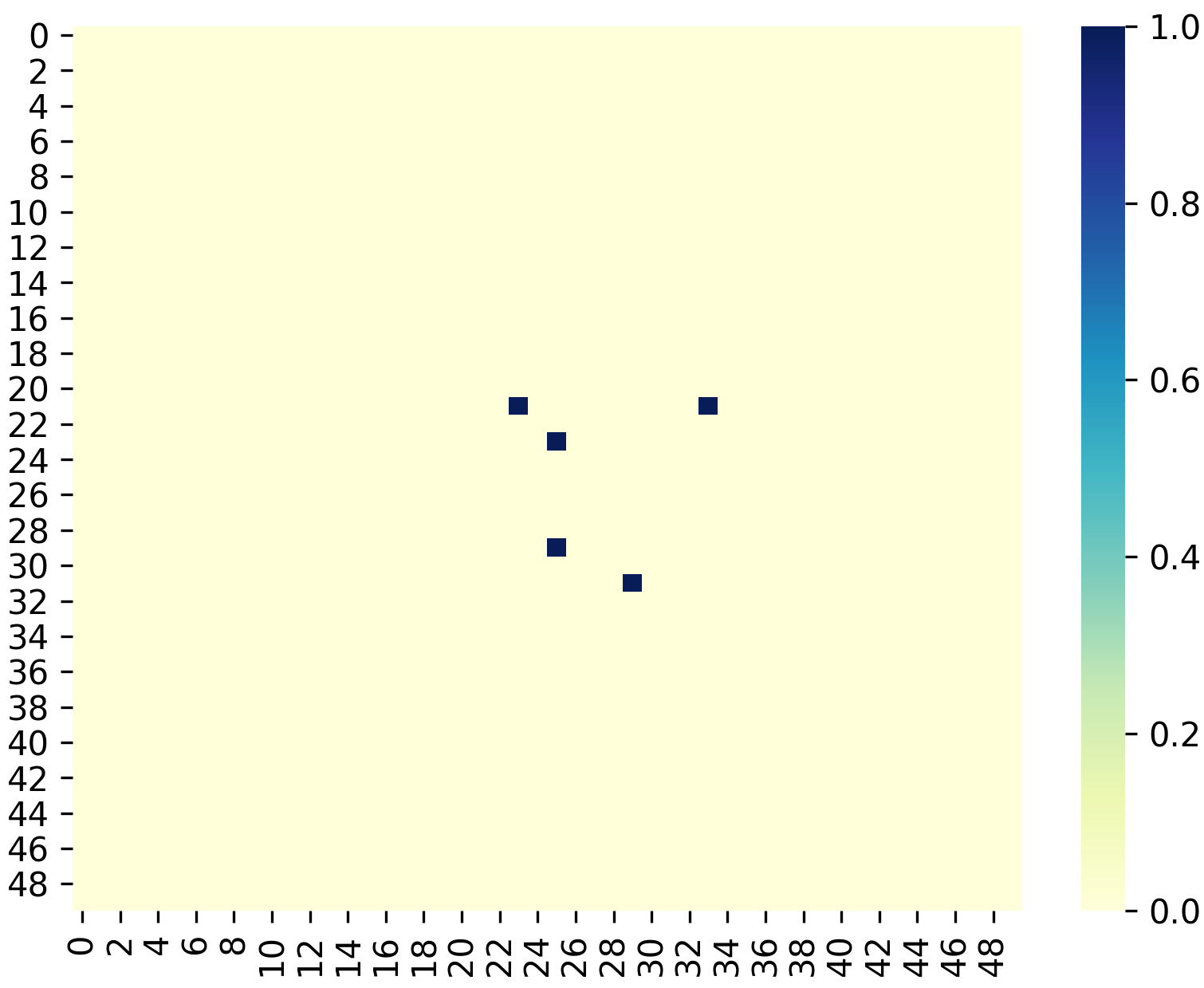}}}
	\subfigure[company]{\label{fig:channel16}\includegraphics[width=0.24\linewidth]{{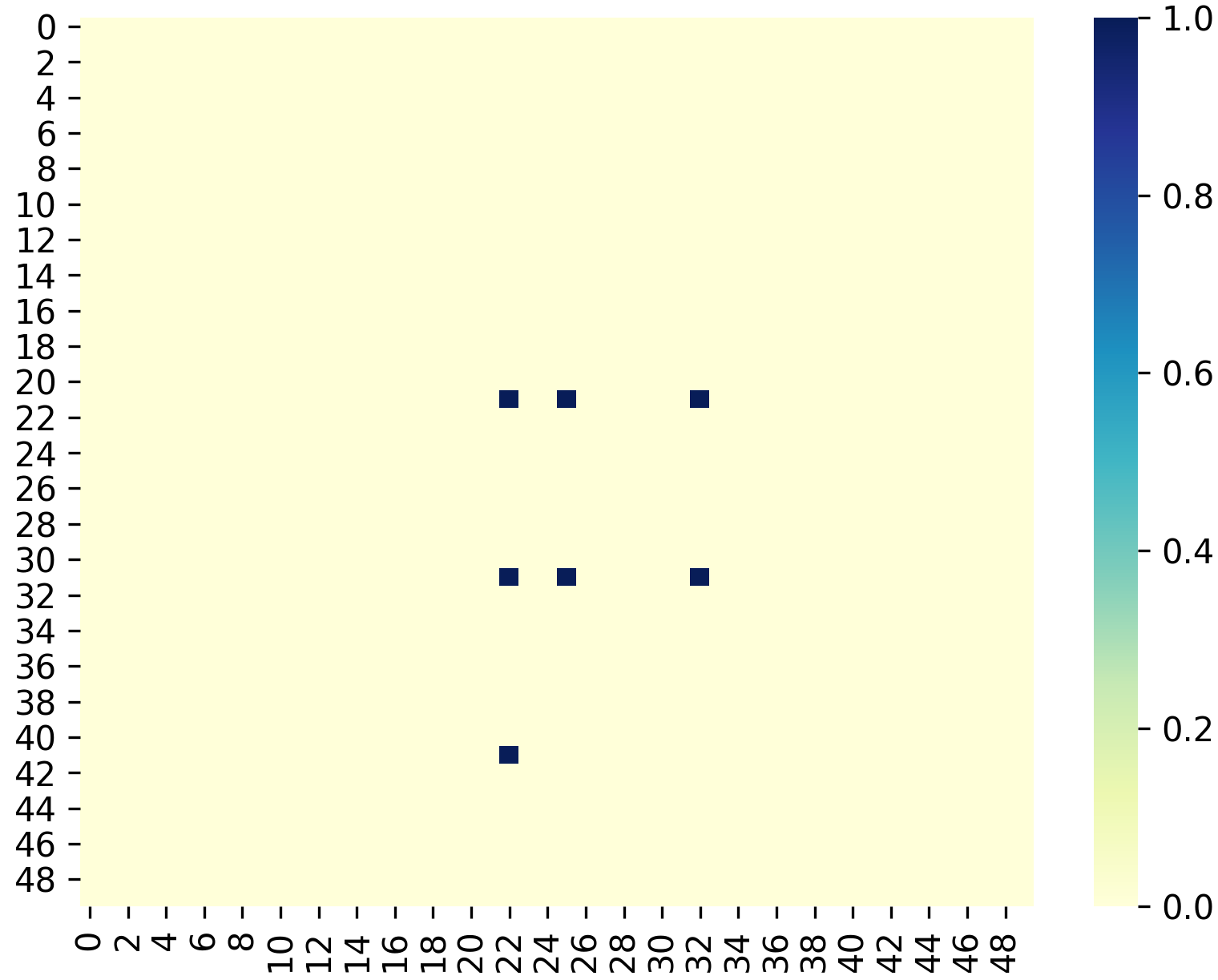}}}
	
	\vspace{-0.1cm}
	
	\caption{Visualization for different POI categories of one generated solution.}
	\label{fig:visual_solution}
	\vspace{-0.6cm}
\end{figure*}

\subsection{Study of The POI Proportion Generated by Different Approaches}

In this experiment, our purpose is to generate a vibrant residential community that owns diverse POI categories and lots of economic activities. 
After obtaining the generated solutions from different generated methods, we count the number of each POI category of different solutions respectively.
Then we visualize the proportion of each POI category of the solutions.

Figure \ref{fig:category_result} shows the comparison of different generated methods. 
In Figure \ref{fig:LUCGAN_poi} , we observe that the LUCGAN generated result owns all POI categories, and 4 (food service), 5 (shopping), 6 (daily life service) and 7 (recreation service) occupy a big proportion in all POI categories. 
In addition, These four POI types are all related to the economic activities closely.
Therefore, the LUCGAN generated solutions satisfy our design scheme.
Figure \ref{fig:vae_poi} shows the POI proportion relation in VAE generated result.
We find that 1 (car service) and 17 (road furniture) are missing, so the POI diversity of VAE is not complete.
Figure \ref{fig:max_poi} is the POI proportion relation in MAX generated result.
Each POI category occupies balanced proportion, which indicates the MAX method is stubborn.
The flexibility of MAX is poorer than VAE and LUCGAN.

\subsection{Study of The Generation for Each POI Category}

We aim to examine the justifiability of the generated configuration for each POI category
To that end, we visualize the POI distribution for each POI category.
Due to the page limitation, we randomly select 12 POIs of a generated solution for visualization
In Figure \ref{fig:visual_solution}, the darker color block represents the larger POI number in the block.
An interesting observation is that the POI distribution of different categories show their unique patterns.
For example, transportation pots are more concentrated, while food service related POIs are more dispersed across the area;
the distribution of car service spots is very similar to the recreation service, the possible reason is that recreation service spots may occupy many parking lots which potentially attract car services.

To sum up, through the above experiments, LUCGAN is capable to generate the land-use configuration solution based on the context environment embedding effectively and flexibly.

\section{Related Work}

\textbf{Representation Learning.} 
The objective of representation learning is to obtain the low-dimensional representation of original data in latent space.
In general, there are three types of representation learning models:
(1) probabilistic graphical models;
(2) manifold learning models;
(3) auto-encoder models.
The probabilistic graphical models build a complex Bayesian network system to learn the representation of uncertain knowledge buried in original data \cite{qiang2019learning}.
The manifold learning models infer low-dimensional manifold of original data based on neighborhood information by non-parametric approaches \cite{zhu2018image}.
The auto-encoder models learn the latent representation by minimizing the reconstruction loss between original and reconstructed data \cite{otto2019linearly}.
In this paper, we utilize the auto-encoder paradigm to learn the representation of the spatial context environment.
There are many successful spatial representation applications with the development of deep representation learning techniques \cite{fu2018representing,wang2018learning,zhang2019unifying,fu2019efficient}.
For instance, Wang et al. capture the feature of GPS trajectory data by utilizing spatio-temporal embedding learning skills, then the embedding is used to analyze driving behavior \cite{wang2019spatiotemporal}.
Du et al. propose a new spatial representation learning framework that captures the static and dynamic characteristics among the spatial entities.
And they utilize the learned spatial representation to improve the performance of real estate price prediction.
\cite{8970913}. 

\textbf{Generative Adversarial Networks.}
Generative Adversarial Networks is a hot research field in recent years \cite{zhang2020curb,zhang2019trafficgan}.
The GAN algorithms can be classified into three categories from task-driven view .
(1). semi-supervised learning GANs. 
Usually, a complete labeled data set is difficult to obtain, the semi-supervised learning GANs can utilize unlabeled data or partially labeled data to train an excellent classifier \cite{ding2018semi,liu2020catgan}. 
For instance, Akcay et al. design a semi-supervised GAN anomaly detection framework that only uses normal data during the training phase \cite{akcay2018ganomaly}.
(2). transfer learning GANs.
Many researchers utilize the transfer learning GANs to transfer knowledge among different domains \cite{hoffman2018cycada,tzeng2017adversarial}.
For instance, Choi et al. build an unified generative adversarial networks to translate the images in different domains \cite{choi2018stargan}.
(3). reinforcement learning GANs.
Generative models are incorporated with reinforcement learning (RL) to improve the model generative performance \cite{sarmad2019rl}.
For instance, Ganin et al. combine reinforce learning and GAN to synthesize high-quality images \cite{ganin2018synthesizing}.

\textbf{Urban Planning.}
Urban planning is a complex research field \cite{adams1994urban}.
The specialists need to consider lots of factors such as government policy, environmental protection to design suitable land-use configuration plan \cite{niemela1999ecology}.
For example, partial researchers focus on constructing an urban planning solution for human health and well-being \cite{barton2013healthy}.
Simultaneously, many researchers form a land-use configuration solution based on the development of real estate \cite{ratcliffe2004urban}.
Therefore, it is difficult to generate a suitable and excellent urban planning solution objectively.
To the best of our knowledge, we first propose an automatic land-use configuration planner to solve this problem.

\section{Conclusion Remarks}
In order to generate a suitable and excellent land-use configuration solution objectively and reduce the heavy burden of urban planning specialists, we proposed an automatic land-use configuration planner framework.
This framework generates the land-use solution based on the context embedding of a virgin area.
Specifically, we obtained the residential community and its context based on the latitude and longitude of residential areas firstly. 
we then extracted the explicit features of the context from three aspects: (1) value-added space;
(2) poi distribution;
(3) traffic condition. 
Afterward, we mapped the explicit feature vectors to the geographical spatial graph as the attributes of the corresponding node.
Next, we utilized the graph embedding technique to fuse all explicit features and spatial relations in the context together to obtain the context embedding.
Then we distinguished the excellent and terrible land-use configuration plans based on expert knowledge.
Finally, the context embedding, excellent and terrible plans were input into our LUCGAN to learn the distribution of excellent plans.
The LUCGAN can generate a suitable and excellent land-use configuration solution based on the context embedding when the model converges.
Ultimately, we conduct extensive experiments to exhibit the effectiveness of our automatic planner.

\section*{Acknowledgment}
This research was partially supported by the National Science Foundation (NSF) via the grant numbers: 1755946, I2040950, 2006889.
\bibliographystyle{ACM-Reference-Format}
\bibliography{acmart}


\begin{thebibliography}{24}


\ifx \showCODEN    \undefined \def \showCODEN     #1{\unskip}     \fi
\ifx \showDOI      \undefined \def \showDOI       #1{#1}\fi
\ifx \showISBNx    \undefined \def \showISBNx     #1{\unskip}     \fi
\ifx \showISBNxiii \undefined \def \showISBNxiii  #1{\unskip}     \fi
\ifx \showISSN     \undefined \def \showISSN      #1{\unskip}     \fi
\ifx \showLCCN     \undefined \def \showLCCN      #1{\unskip}     \fi
\ifx \shownote     \undefined \def \shownote      #1{#1}          \fi
\ifx \showarticletitle \undefined \def \showarticletitle #1{#1}   \fi
\ifx \showURL      \undefined \def \showURL       {\relax}        \fi
\providecommand\bibfield[2]{#2}
\providecommand\bibinfo[2]{#2}
\providecommand\natexlab[1]{#1}
\providecommand\showeprint[2][]{arXiv:#2}

\bibitem[\protect\citeauthoryear{Adams}{Adams}{1994}]%
        {adams1994urban}
\bibfield{author}{\bibinfo{person}{David Adams}.}
  \bibinfo{year}{1994}\natexlab{}.
\newblock \bibinfo{booktitle}{\emph{Urban planning and the development
  process}}.
\newblock \bibinfo{publisher}{Psychology Press}.
\newblock


\bibitem[\protect\citeauthoryear{Akcay, Atapour-Abarghouei, and Breckon}{Akcay
  et~al\mbox{.}}{2018}]%
        {akcay2018ganomaly}
\bibfield{author}{\bibinfo{person}{Samet Akcay}, \bibinfo{person}{Amir
  Atapour-Abarghouei}, {and} \bibinfo{person}{Toby~P Breckon}.}
  \bibinfo{year}{2018}\natexlab{}.
\newblock \showarticletitle{Ganomaly: Semi-supervised anomaly detection via
  adversarial training}. In \bibinfo{booktitle}{\emph{Asian conference on
  computer vision}}. Springer, \bibinfo{pages}{622--637}.
\newblock


\bibitem[\protect\citeauthoryear{Barton and Tsourou}{Barton and
  Tsourou}{2013}]%
        {barton2013healthy}
\bibfield{author}{\bibinfo{person}{Hugh Barton} {and}
  \bibinfo{person}{Catherine Tsourou}.} \bibinfo{year}{2013}\natexlab{}.
\newblock \bibinfo{booktitle}{\emph{Healthy urban planning}}.
\newblock \bibinfo{publisher}{Routledge}.
\newblock


\bibitem[\protect\citeauthoryear{Choi, Choi, Kim, Ha, Kim, and Choo}{Choi
  et~al\mbox{.}}{2018}]%
        {choi2018stargan}
\bibfield{author}{\bibinfo{person}{Yunjey Choi}, \bibinfo{person}{Minje Choi},
  \bibinfo{person}{Munyoung Kim}, \bibinfo{person}{Jung-Woo Ha},
  \bibinfo{person}{Sunghun Kim}, {and} \bibinfo{person}{Jaegul Choo}.}
  \bibinfo{year}{2018}\natexlab{}.
\newblock \showarticletitle{Stargan: Unified generative adversarial networks
  for multi-domain image-to-image translation}. In
  \bibinfo{booktitle}{\emph{Proceedings of the IEEE conference on computer
  vision and pattern recognition}}. \bibinfo{pages}{8789--8797}.
\newblock


\bibitem[\protect\citeauthoryear{Ding, Tang, and Zhang}{Ding
  et~al\mbox{.}}{2018}]%
        {ding2018semi}
\bibfield{author}{\bibinfo{person}{Ming Ding}, \bibinfo{person}{Jie Tang},
  {and} \bibinfo{person}{Jie Zhang}.} \bibinfo{year}{2018}\natexlab{}.
\newblock \showarticletitle{Semi-supervised learning on graphs with generative
  adversarial nets}. In \bibinfo{booktitle}{\emph{Proceedings of the 27th ACM
  International Conference on Information and Knowledge Management}}.
  \bibinfo{pages}{913--922}.
\newblock


\bibitem[\protect\citeauthoryear{{Du}, {Zhang}, {Wang}, {Leopold}, and
  {Fu}}{{Du} et~al\mbox{.}}{2019}]%
        {8970913}
\bibfield{author}{\bibinfo{person}{J. {Du}}, \bibinfo{person}{Y. {Zhang}},
  \bibinfo{person}{P. {Wang}}, \bibinfo{person}{J. {Leopold}}, {and}
  \bibinfo{person}{Y. {Fu}}.} \bibinfo{year}{2019}\natexlab{}.
\newblock \showarticletitle{Beyond Geo-First Law: Learning Spatial
  Representations via Integrated Autocorrelations and Complementarity}. In
  \bibinfo{booktitle}{\emph{2019 IEEE International Conference on Data Mining
  (ICDM)}}. \bibinfo{pages}{160--169}.
\newblock


\bibitem[\protect\citeauthoryear{Fu, Liu, Ge, Wang, Zhu, Li, and Xiong}{Fu
  et~al\mbox{.}}{2018}]%
        {fu2018representing}
\bibfield{author}{\bibinfo{person}{Yanjie Fu}, \bibinfo{person}{Guannan Liu},
  \bibinfo{person}{Yong Ge}, \bibinfo{person}{Pengyang Wang},
  \bibinfo{person}{Hengshu Zhu}, \bibinfo{person}{Chunxiao Li}, {and}
  \bibinfo{person}{Hui Xiong}.} \bibinfo{year}{2018}\natexlab{}.
\newblock \showarticletitle{Representing urban forms: A collective learning
  model with heterogeneous human mobility data}.
\newblock \bibinfo{journal}{\emph{IEEE transactions on knowledge and data
  engineering}} \bibinfo{volume}{31}, \bibinfo{number}{3}
  (\bibinfo{year}{2018}), \bibinfo{pages}{535--548}.
\newblock


\bibitem[\protect\citeauthoryear{Fu, Wang, Du, Wu, and Li}{Fu
  et~al\mbox{.}}{2019}]%
        {fu2019efficient}
\bibfield{author}{\bibinfo{person}{Yanjie Fu}, \bibinfo{person}{Pengyang Wang},
  \bibinfo{person}{Jiadi Du}, \bibinfo{person}{Le Wu}, {and}
  \bibinfo{person}{Xiaolin Li}.} \bibinfo{year}{2019}\natexlab{}.
\newblock \showarticletitle{Efficient region embedding with multi-view spatial
  networks: A perspective of locality-constrained spatial autocorrelations}. In
  \bibinfo{booktitle}{\emph{Proceedings of the AAAI Conference on Artificial
  Intelligence}}, Vol.~\bibinfo{volume}{33}. \bibinfo{pages}{906--913}.
\newblock


\bibitem[\protect\citeauthoryear{Ganin, Kulkarni, Babuschkin, Eslami, and
  Vinyals}{Ganin et~al\mbox{.}}{2018}]%
        {ganin2018synthesizing}
\bibfield{author}{\bibinfo{person}{Yaroslav Ganin}, \bibinfo{person}{Tejas
  Kulkarni}, \bibinfo{person}{Igor Babuschkin}, \bibinfo{person}{SM~Ali
  Eslami}, {and} \bibinfo{person}{Oriol Vinyals}.}
  \bibinfo{year}{2018}\natexlab{}.
\newblock \showarticletitle{Synthesizing Programs for Images using Reinforced
  Adversarial Learning}. In \bibinfo{booktitle}{\emph{International Conference
  on Machine Learning}}. \bibinfo{pages}{1666--1675}.
\newblock


\bibitem[\protect\citeauthoryear{Hoffman, Tzeng, Park, Zhu, Isola, Saenko,
  Efros, and Darrell}{Hoffman et~al\mbox{.}}{2018}]%
        {hoffman2018cycada}
\bibfield{author}{\bibinfo{person}{Judy Hoffman}, \bibinfo{person}{Eric Tzeng},
  \bibinfo{person}{Taesung Park}, \bibinfo{person}{Jun-Yan Zhu},
  \bibinfo{person}{Phillip Isola}, \bibinfo{person}{Kate Saenko},
  \bibinfo{person}{Alexei Efros}, {and} \bibinfo{person}{Trevor Darrell}.}
  \bibinfo{year}{2018}\natexlab{}.
\newblock \showarticletitle{CyCADA: Cycle-Consistent Adversarial Domain
  Adaptation}. In \bibinfo{booktitle}{\emph{International Conference on Machine
  Learning}}. \bibinfo{pages}{1989--1998}.
\newblock


\bibitem[\protect\citeauthoryear{Liu, Wang, and Liang}{Liu
  et~al\mbox{.}}{2020}]%
        {liu2020catgan}
\bibfield{author}{\bibinfo{person}{Zhiyue Liu}, \bibinfo{person}{Jiahai Wang},
  {and} \bibinfo{person}{Zhiwei Liang}.} \bibinfo{year}{2020}\natexlab{}.
\newblock \showarticletitle{CatGAN: Category-Aware Generative Adversarial
  Networks with Hierarchical Evolutionary Learning for Category Text
  Generation.}. In \bibinfo{booktitle}{\emph{AAAI}}.
  \bibinfo{pages}{8425--8432}.
\newblock


\bibitem[\protect\citeauthoryear{Niemel{\"a}}{Niemel{\"a}}{1999}]%
        {niemela1999ecology}
\bibfield{author}{\bibinfo{person}{Jari Niemel{\"a}}.}
  \bibinfo{year}{1999}\natexlab{}.
\newblock \showarticletitle{Ecology and urban planning}.
\newblock \bibinfo{journal}{\emph{Biodiversity \& Conservation}}
  \bibinfo{volume}{8}, \bibinfo{number}{1} (\bibinfo{year}{1999}),
  \bibinfo{pages}{119--131}.
\newblock


\bibitem[\protect\citeauthoryear{Otto and Rowley}{Otto and Rowley}{2019}]%
        {otto2019linearly}
\bibfield{author}{\bibinfo{person}{Samuel~E Otto} {and}
  \bibinfo{person}{Clarence~W Rowley}.} \bibinfo{year}{2019}\natexlab{}.
\newblock \showarticletitle{Linearly recurrent autoencoder networks for
  learning dynamics}.
\newblock \bibinfo{journal}{\emph{SIAM Journal on Applied Dynamical Systems}}
  \bibinfo{volume}{18}, \bibinfo{number}{1} (\bibinfo{year}{2019}),
  \bibinfo{pages}{558--593}.
\newblock


\bibitem[\protect\citeauthoryear{Qiang, Fu, Yu, Guo, Zhou, and Sigal}{Qiang
  et~al\mbox{.}}{2019}]%
        {qiang2019learning}
\bibfield{author}{\bibinfo{person}{Yu-Ting Qiang}, \bibinfo{person}{Yan-Wei
  Fu}, \bibinfo{person}{Xiao Yu}, \bibinfo{person}{Yan-Wen Guo},
  \bibinfo{person}{Zhi-Hua Zhou}, {and} \bibinfo{person}{Leonid Sigal}.}
  \bibinfo{year}{2019}\natexlab{}.
\newblock \showarticletitle{Learning to generate posters of scientific papers
  by probabilistic graphical models}.
\newblock \bibinfo{journal}{\emph{Journal of Computer Science and Technology}}
  \bibinfo{volume}{34}, \bibinfo{number}{1} (\bibinfo{year}{2019}),
  \bibinfo{pages}{155--169}.
\newblock


\bibitem[\protect\citeauthoryear{Ratcliffe, Stubbs, and Shepherd}{Ratcliffe
  et~al\mbox{.}}{2004}]%
        {ratcliffe2004urban}
\bibfield{author}{\bibinfo{person}{John Ratcliffe}, \bibinfo{person}{Michael
  Stubbs}, {and} \bibinfo{person}{Mark Shepherd}.}
  \bibinfo{year}{2004}\natexlab{}.
\newblock \bibinfo{booktitle}{\emph{Urban planning and real estate
  development}}. Vol.~\bibinfo{volume}{8}.
\newblock \bibinfo{publisher}{Taylor \& Francis}.
\newblock


\bibitem[\protect\citeauthoryear{Sarmad, Lee, and Kim}{Sarmad
  et~al\mbox{.}}{2019}]%
        {sarmad2019rl}
\bibfield{author}{\bibinfo{person}{Muhammad Sarmad},
  \bibinfo{person}{Hyunjoo~Jenny Lee}, {and} \bibinfo{person}{Young~Min Kim}.}
  \bibinfo{year}{2019}\natexlab{}.
\newblock \showarticletitle{Rl-gan-net: A reinforcement learning agent
  controlled gan network for real-time point cloud shape completion}. In
  \bibinfo{booktitle}{\emph{Proceedings of the IEEE Conference on Computer
  Vision and Pattern Recognition}}. \bibinfo{pages}{5898--5907}.
\newblock


\bibitem[\protect\citeauthoryear{Tzeng, Hoffman, Saenko, and Darrell}{Tzeng
  et~al\mbox{.}}{2017}]%
        {tzeng2017adversarial}
\bibfield{author}{\bibinfo{person}{Eric Tzeng}, \bibinfo{person}{Judy Hoffman},
  \bibinfo{person}{Kate Saenko}, {and} \bibinfo{person}{Trevor Darrell}.}
  \bibinfo{year}{2017}\natexlab{}.
\newblock \showarticletitle{Adversarial discriminative domain adaptation}. In
  \bibinfo{booktitle}{\emph{Proceedings of the IEEE conference on computer
  vision and pattern recognition}}. \bibinfo{pages}{7167--7176}.
\newblock


\bibitem[\protect\citeauthoryear{Wang, Fu, Zhang, Li, and Lin}{Wang
  et~al\mbox{.}}{2018a}]%
        {wang2018learning}
\bibfield{author}{\bibinfo{person}{Pengyang Wang}, \bibinfo{person}{Yanjie Fu},
  \bibinfo{person}{Jiawei Zhang}, \bibinfo{person}{Xiaolin Li}, {and}
  \bibinfo{person}{Dan Lin}.} \bibinfo{year}{2018}\natexlab{a}.
\newblock \showarticletitle{Learning urban community structures: A collective
  embedding perspective with periodic spatial-temporal mobility graphs}.
\newblock \bibinfo{journal}{\emph{ACM Transactions on Intelligent Systems and
  Technology (TIST)}} \bibinfo{volume}{9}, \bibinfo{number}{6}
  (\bibinfo{year}{2018}), \bibinfo{pages}{1--28}.
\newblock


\bibitem[\protect\citeauthoryear{Wang, Li, Zheng, Aggarwal, and Fu}{Wang
  et~al\mbox{.}}{2019}]%
        {wang2019spatiotemporal}
\bibfield{author}{\bibinfo{person}{Pengyang Wang}, \bibinfo{person}{Xiaolin
  Li}, \bibinfo{person}{Yu Zheng}, \bibinfo{person}{Charu Aggarwal}, {and}
  \bibinfo{person}{Yanjie Fu}.} \bibinfo{year}{2019}\natexlab{}.
\newblock \showarticletitle{Spatiotemporal Representation Learning for Driving
  Behavior Analysis: A Joint Perspective of Peer and Temporal Dependencies}.
\newblock \bibinfo{journal}{\emph{IEEE Transactions on Knowledge and Data
  Engineering}} (\bibinfo{year}{2019}).
\newblock


\bibitem[\protect\citeauthoryear{Wang, Zhang, Liu, Fu, and Aggarwal}{Wang
  et~al\mbox{.}}{2018b}]%
        {wang2018ensemble}
\bibfield{author}{\bibinfo{person}{Pengyang Wang}, \bibinfo{person}{Jiawei
  Zhang}, \bibinfo{person}{Guannan Liu}, \bibinfo{person}{Yanjie Fu}, {and}
  \bibinfo{person}{Charu Aggarwal}.} \bibinfo{year}{2018}\natexlab{b}.
\newblock \showarticletitle{Ensemble-spotting: Ranking urban vibrancy via poi
  embedding with multi-view spatial graphs}. In
  \bibinfo{booktitle}{\emph{Proceedings of the 2018 SIAM International
  Conference on Data Mining}}. SIAM, \bibinfo{pages}{351--359}.
\newblock


\bibitem[\protect\citeauthoryear{Zhang, Fu, Wang, Li, and Zheng}{Zhang
  et~al\mbox{.}}{2019a}]%
        {zhang2019unifying}
\bibfield{author}{\bibinfo{person}{Yunchao Zhang}, \bibinfo{person}{Yanjie Fu},
  \bibinfo{person}{Pengyang Wang}, \bibinfo{person}{Xiaolin Li}, {and}
  \bibinfo{person}{Yu Zheng}.} \bibinfo{year}{2019}\natexlab{a}.
\newblock \showarticletitle{Unifying inter-region autocorrelation and
  intra-region structures for spatial embedding via collective adversarial
  learning}. In \bibinfo{booktitle}{\emph{Proceedings of the 25th ACM SIGKDD
  International Conference on Knowledge Discovery \& Data Mining}}.
  \bibinfo{pages}{1700--1708}.
\newblock


\bibitem[\protect\citeauthoryear{Zhang, Li, Zhou, Kong, and Luo}{Zhang
  et~al\mbox{.}}{2019b}]%
        {zhang2019trafficgan}
\bibfield{author}{\bibinfo{person}{Yingxue Zhang}, \bibinfo{person}{Yanhua Li},
  \bibinfo{person}{Xun Zhou}, \bibinfo{person}{Xiangnan Kong}, {and}
  \bibinfo{person}{Jun Luo}.} \bibinfo{year}{2019}\natexlab{b}.
\newblock \showarticletitle{TrafficGAN: Off-Deployment Traffic Estimation with
  Traffic Generative Adversarial Networks}. In \bibinfo{booktitle}{\emph{2019
  IEEE International Conference on Data Mining (ICDM)}}. IEEE,
  \bibinfo{pages}{1474--1479}.
\newblock


\bibitem[\protect\citeauthoryear{Zhang, Li, Zhou, Kong, and Luo}{Zhang
  et~al\mbox{.}}{2020}]%
        {zhang2020curb}
\bibfield{author}{\bibinfo{person}{Yingxue Zhang}, \bibinfo{person}{Yanhua Li},
  \bibinfo{person}{Xun Zhou}, \bibinfo{person}{Xiangnan Kong}, {and}
  \bibinfo{person}{Jun Luo}.} \bibinfo{year}{2020}\natexlab{}.
\newblock \showarticletitle{Curb-GAN: Conditional Urban Traffic Estimation
  through Spatio-Temporal Generative Adversarial Networks}. In
  \bibinfo{booktitle}{\emph{Proceedings of the 26th ACM SIGKDD International
  Conference on Knowledge Discovery \& Data Mining}}.
  \bibinfo{pages}{842--852}.
\newblock


\bibitem[\protect\citeauthoryear{Zhu, Liu, Cauley, Rosen, and Rosen}{Zhu
  et~al\mbox{.}}{2018}]%
        {zhu2018image}
\bibfield{author}{\bibinfo{person}{Bo Zhu}, \bibinfo{person}{Jeremiah~Z Liu},
  \bibinfo{person}{Stephen~F Cauley}, \bibinfo{person}{Bruce~R Rosen}, {and}
  \bibinfo{person}{Matthew~S Rosen}.} \bibinfo{year}{2018}\natexlab{}.
\newblock \showarticletitle{Image reconstruction by domain-transform manifold
  learning}.
\newblock \bibinfo{journal}{\emph{Nature}} \bibinfo{volume}{555},
  \bibinfo{number}{7697} (\bibinfo{year}{2018}), \bibinfo{pages}{487--492}.
\newblock


\end{thebibliography}

\end{document}